\pdfoutput=1

\documentclass[11pt]{article}

\usepackage[final]{acl}
\usepackage{times}
\usepackage{latexsym}
\usepackage{amsmath}
\usepackage{amssymb}
\usepackage{bm}
\usepackage{bbm}
\usepackage[ruled,vlined,linesnumbered]{algorithm2e}
\usepackage{setspace}                                
\usepackage{graphicx}   
\usepackage{multirow, array}
\usepackage{booktabs}
\usepackage{color}
\usepackage{arydshln}
\usepackage{tcolorbox}
\tcbuselibrary{listings,breakable}
\tcbuselibrary{skins} 
\usepackage{colortbl}
\usepackage{xcolor}
\usepackage{pifont}
\definecolor{RedOrange}{RGB}{255,69,0}
\definecolor{BlueGreen}{RGB}{0,128,128}
\definecolor{bittersweet}{rgb}{1.0, 0.44, 0.37}
\definecolor{mygreen}{rgb}{0.29, 0.7, 0.48}
\usepackage{pifont}

\definecolor{demphcolor}{RGB}{144,144,144}

\definecolor{mygray}{gray}{0.4}
\definecolor{autopurple}{HTML}{7030A0}
\definecolor{dyna_yellow}{HTML}{BF9000}
\definecolor{adaptive_blue}{HTML}{0070C0}
\definecolor{darksalmon}{rgb}{0.91, 0.59, 0.48}
\definecolor{emerald}{rgb}{0.31, 0.78, 0.47}
\definecolor{green(pigment)}{rgb}{0.0, 0.65, 0.31}
\definecolor{amaranth}{rgb}{0.9, 0.17, 0.31}
\definecolor{iris}{rgb}{0.35, 0.31, 0.81}
\definecolor{uu}{rgb}{0.95, 0.51, 0.51}
\definecolor{spirodiscoball}{rgb}{0.06, 0.75, 0.99}
\definecolor{myforestgreen}{RGB}{34,139,34}

\definecolor{softblue}{RGB}{121,151,161}
\definecolor{softgreen}{RGB}{106,176,106}
\definecolor{softpink}{RGB}{179,127,135}
\definecolor{softpurple}{RGB}{151,134,151}
\definecolor{softbrown}{RGB}{147,126,98}
\definecolor{softorange}{RGB}{232,162,104}    
\definecolor{softyellow}{RGB}{232,216,120}   
\definecolor{softred}{RGB}{210,120,120}       
\definecolor{softteal}{RGB}{120,180,175}      
\definecolor{softgray}{RGB}{180,180,180}    
\definecolor{softcyan}{RGB}{140,200,200}     
\definecolor{softlavender}{RGB}{200,180,220}  
\definecolor{softpeach}{RGB}{245,200,170}

\usepackage{cleveref}
\SetKwInOut{Input}{Input}\SetKwInOut{Output}{Output}

\SetCommentSty{mycommfont}
\SetKwComment{Comment}{$\triangleright$\ }{}

\usepackage{xspace}

\definecolor{ada_blue}{rgb}{0,205,205}
\definecolor{glt_red}{rgb}{109,205,255}
\definecolor{MorandiBlue}{RGB}{118,134,146}

\definecolor{demphcolor}{RGB}{144,144,144}
\definecolor{mygray}{gray}{0.4}
\definecolor{autopurple}{HTML}{7030A0}
\definecolor{dyna_yellow}{HTML}{BF9000}
\definecolor{adaptive_blue}{HTML}{0070C0}
\definecolor{darkgrey}{RGB}{120,120,120}
\definecolor{mygrey}{RGB}{200,200,200}

\newcommand{\Checkmark}{\ding{51}}

\usepackage[font=small,labelfont=bf]{caption} 
\usepackage{makecell}
\usepackage{tabulary}

\definecolor{myblue}{HTML}{00CDCD}
\definecolor{champagne}{rgb}{0.74, 0.83, 0.9}
\definecolor{champagne}{rgb}{0.97, 0.91, 0.81}

\usepackage[T1]{fontenc}

\usepackage[utf8]{inputenc}

\usepackage[nopatch=footnote]{microtype}

\usepackage{inconsolata}

\usepackage{graphicx}

\title{AgentInit: Initializing LLM-based Multi-Agent Systems via Diversity and Expertise Orchestration for Effective and Efficient Collaboration}

\author{Chunhao Tian$^{1}$~~
        Yutong Wang$^{1}$~~
        Xuebo Liu$^{1}$\thanks{~~Corresponding Author}~~ \\
        \textbf{Zhexuan Wang}$^{1}$~~ 
        \textbf{Liang Ding}$^{2}$~~ 
        \textbf{Miao Zhang}$^{3}$~~
        \textbf{Min Zhang}$^{1}$ \\
        $^{1}$Institute of Computing and Intelligence, Harbin Institute of Technology, Shenzhen, China \\
        $^{2}$The University of Sydney, Sydney, Australia \\
        $^{3}$School of Computer Science and Technology, Harbin Institute of Technology, Shenzhen, China \\
        \texttt{\{tianchunhao,wangyutong,wangzhexuan\}@stu.hit.edu.cn, liangding.liam@gmail.com} \\
        \texttt{\{liuxuebo,zhangmiao,zhangmin2021\}@hit.edu.cn}
  }

\begin{document}
\maketitle
\begin{abstract}
Proper initialization is crucial for any system, particularly in multi-agent systems (MAS), where it plays a pivotal role in determining both the system's efficiency and effectiveness.
However, existing MAS initialization methods do not fully account for the collaborative needs of the generated agents in subsequent stages.
Inspired by the principles of effective team composition, we propose \textbf{AgentInit}, which aims to optimize the structure of agent teams. Specifically, in addition to multi-round interactions and reflections between agents during agent generation, AgentInit incorporates a Natural Language to Format mechanism to ensure consistency and standardization. 
Balanced team selection strategies using Pareto principles are subsequently applied to jointly consider agent team diversity and task relevance to promote effective and efficient collaboration and enhance overall system performance.
Experiments show that AgentInit consistently outperforms state-of-the-art initialization methods and pre-defined strategies across various frameworks and tasks, achieving an overall performance improvement of up to 1.2 and 1.6, respectively, while also significantly reducing token consumption.
Further analysis confirms its strong transferability to similar tasks and verifies the effectiveness of its key components, demonstrating its capability and adaptability as a reliable MAS initialization method.
Source code and models are available at \url{https://github.com/1737423697/AgentInit}.
\end{abstract}

\section{Introduction}
Recent advances in Large Language Models~(LLMs) \citep{team2024gemini,grattafiori2024llama3herdmodels,openai2024gpt4technicalreport} have enabled autonomous agents \citep{yao2023react, Wang_2024} capable of complex task solving \citep{wang2025delta}. Extending beyond single-agent settings, Multi-Agent Systems~(MAS) provide a promising framework for coordinating diverse agents toward human-like collaboration and improved performance \citep{hao2023chatllmnetworkbrainsintelligence,liang2024encouraging,du2024improving,zhuge2024gptswarm,wang2025mixtureofagents,song2025adaptiveinconversationteambuilding}.
Initialization plays a critical role in MAS, as it defines the roles and responsibilities of each agent, thereby ensuring efficient task execution and enhancing overall system performance \citep{suzgun2024metapromptingenhancinglanguagemodels}. While many existing MAS frameworks still rely on manual design \citep{wu2023autogenenablingnextgenllm,li2023camel,hong2024metagpt}, recent approaches such as AgentVerse \citep{chen2023agentversefacilitatingmultiagentcollaboration}, AutoAgents \citep{chen2024autoagents}, and EvoAgent \citep{yuan2025evoagentautomaticmultiagentgeneration} aim to automate agent generation. This automation represents a key step in the initialization process and contributes to improved scalability and adaptability of MAS.

\begin{figure}
    \centering
    \includegraphics[width=0.9\linewidth]{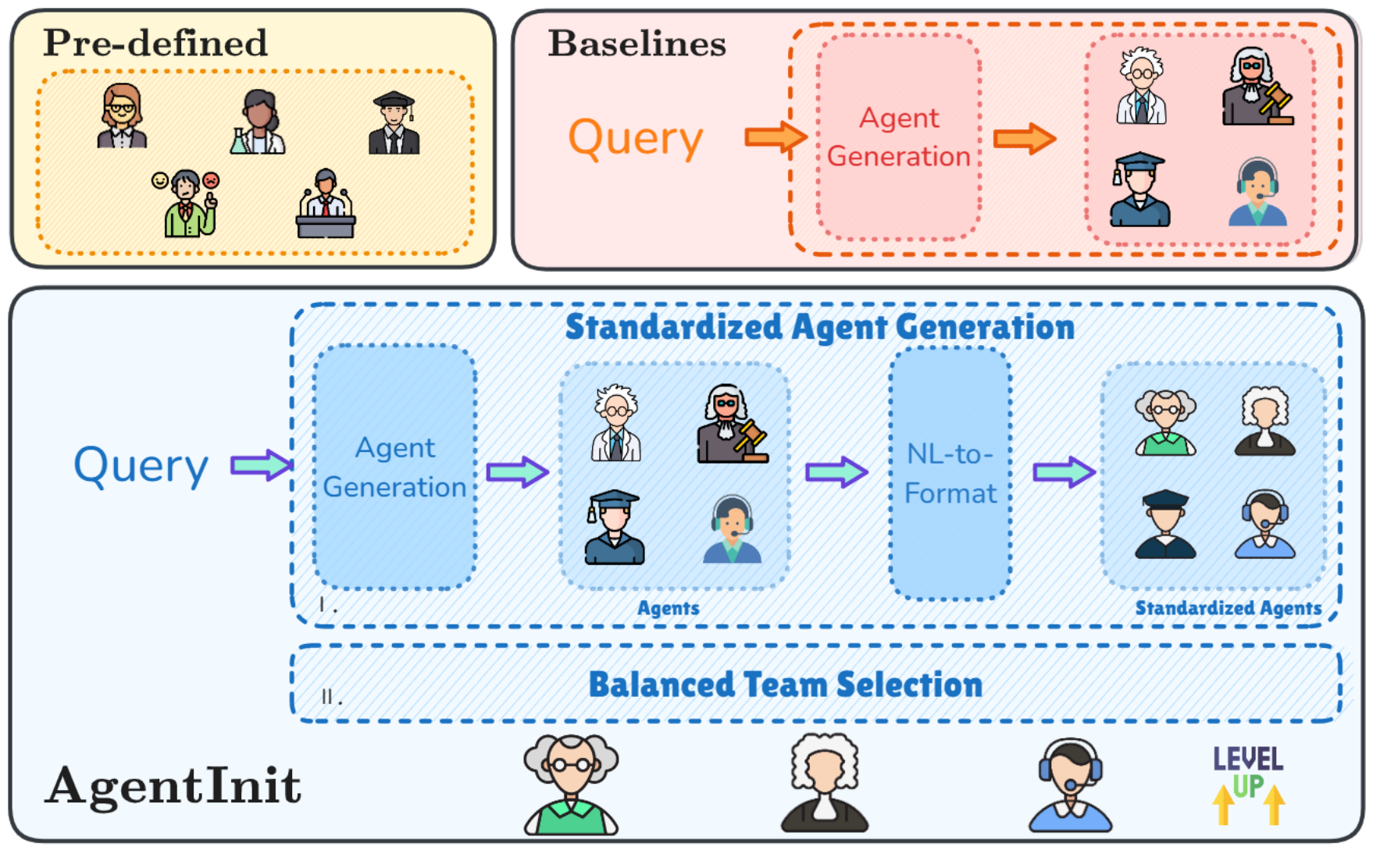}
    \caption{Overview of AgentInit compared with predefined methods and baselines. While the baselines directly generate all agents, AgentInit applies an additional selection module after generation to improve team composition.}
    \label{fig:Intro}
\end{figure}
While automated MAS initialization has made progress, existing methods still struggle to adequately consider effective future collaboration between agents. This leads to teams with irrelevant and overlapping agents, which can cause task derailment and step repetition \citep{cemri2025multiagentllmsystemsfail}, negatively impacting overall system performance.
Existing initialization methods attempt to optimize team configuration using LLMs, but these approaches are often constrained by inherent biases \citep{zheng2023judgingllmasajudgemtbenchchatbot,wang2023largelanguagemodelsfair}. For example, LLMs tend to avoid criticizing their own outputs due to self-preference bias~\citep{wataoka2024selfpreferencebiasllmasajudge}, which limits their ability to identify and eliminate redundant or low-quality agents. As a result, these methods struggle to provide an efficient and well-coordinated team structure at the initialization stage, highlighting the need for a more robust initialization mechanism.

The challenges of initializing MAS suggest the need for structural optimization. In real-world organizations, streamlining personnel and refining team structures improve efficiency~\citep{kozlowski2006enhancing}. Similarly, recent MAS research shows that removing redundant agents enhances system performance~\citep{wang2025agentdropoutdynamicagentelimination}.
In light of these findings, we propose AgentInit, a framework that optimizes team structure by balancing diversity and task relevance. As illustrated in Figure~\ref{fig:Intro}, AgentInit consists of: (1) \textbf{Standardized Agent Generation}:
It decomposes the query into sub-tasks and generates a set of candidate agents over multiple iterations, with feedback used to refine them. To enhance agent quality and ensure a fairer selection process, we apply the NL-to-Format approach~\citep{tam2024letspeakfreelystudy}, which standardizes agent representations for more consistent evaluation.
 (2) \textbf{Balanced Team Selection}: This module treats team optimization as a multi-objective problem, considering two main objectives: task relevance and agent diversity. By constructing a Pareto optimal set~\citep{coello2006evolutionary,deb2011multi}, the LLM-powered selector identifies the most effective team compositions.

A series of experiments is conducted across interactive scientific simulations and diverse NLP tasks, such as reasoning, mathematics, code generation, and writing, demonstrating that AgentInit consistently outperforms existing initialization methods in terms of performance. Additionally, experiments across various MAS frameworks further validate its superior adaptability.

In summary, this paper makes the following contributions:
\begin{itemize}
    \item We propose \textbf{AgentInit}, an effective and efficient MAS initialization method consisting of two modules: \textbf{Standardized Agent Generation} to create capable and standardized agents, and \textbf{Balanced Team Selection} to optimize team composition.
    \item We validate that AgentInit consistently outperforms existing methods, achieving significant improvements in both performance and efficiency through experiments across a wide range of tasks and MAS frameworks.
    \item We explore evaluation objectives for agent selection, identifying task relevance and intra-team diversity as key indicators to guide MAS initialization. 
\end{itemize}
\section{Preliminary}
\begin{figure*}[t]
\centering 
\includegraphics[height=7.3cm]{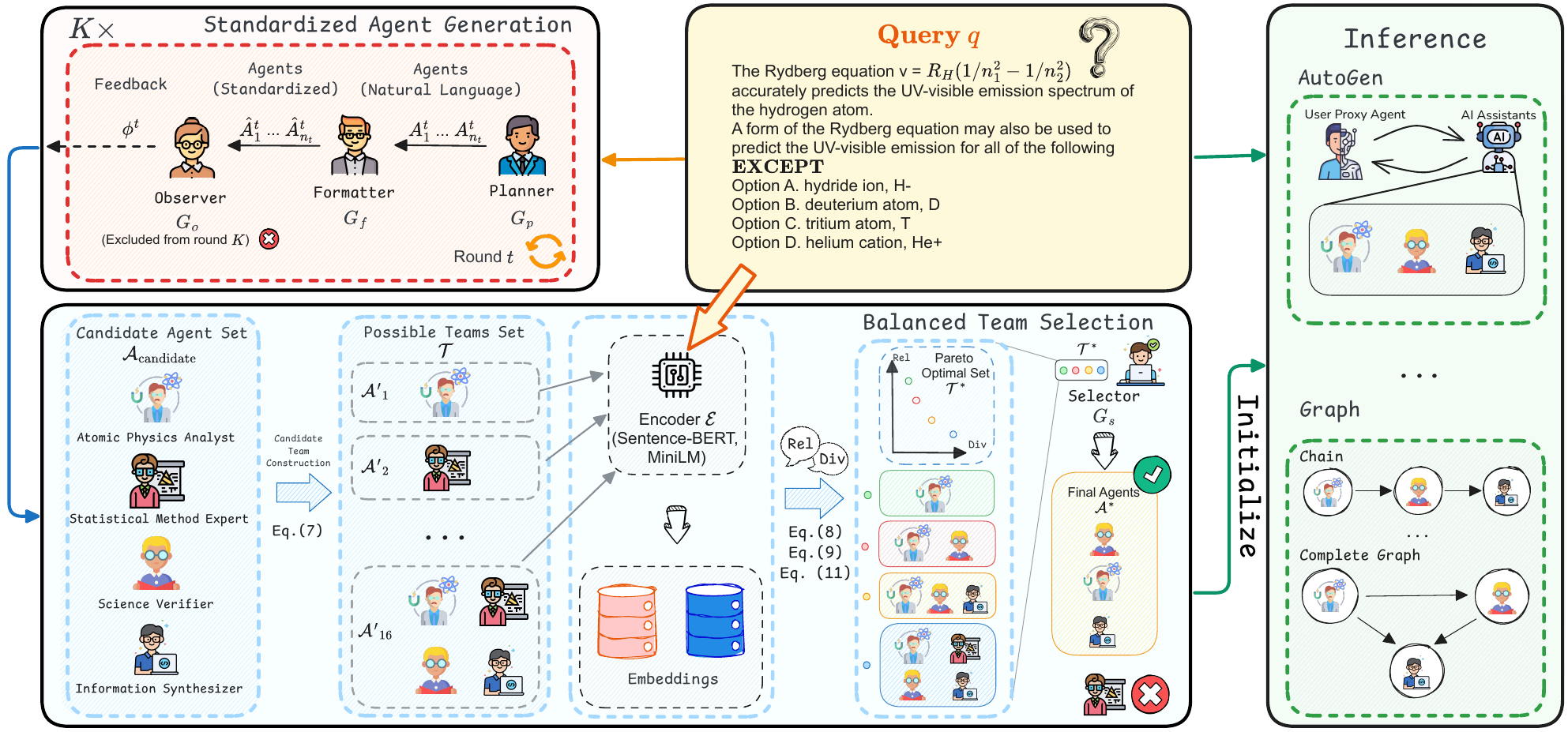}

\caption{The overall process of AgentInit, where the red section involves multi-round iterations to generate standardized agents, the blue section selects the most suitable agent team based on diversity and relevance, and the green section shows AgentInit's application across different inference frameworks.}
\label{fig:MAIN}
\end{figure*}

\subsection{LLM-Based Multi-Agent Systems}\label{sec:MAS bg}

LLM-based MAS have emerged as a key paradigm for collaborative reasoning in complex tasks. Such systems can be defined as \( \mathcal{S} = \{ \mathcal{A}, F \} \), where \( \mathcal{A} = \{ A_1, A_2, \dots, A_n \} \) is the set of agents, and \( F \) is the framework governing their interactions.

The framework \( F \) can be classified into two main categories. The \textbf{Loosely Modeled} frameworks, such as AutoGen~\citep{wu2023autogenenablingnextgenllm} and CAMEL~\citep{li2023camel}, facilitate agent interactions through dynamic, dialogue-driven processes without imposing strict structural constraints.
On the other hand, the \textbf{Formally Modeled} frameworks define agent interactions through formal structures, such as neural networks~\citep{liu2023dynamic}, graphs~\citep{zhuge2024gptswarm,zhang2025cut}, and code~\citep{hu2025automateddesignagenticsystems,zhang2025aflowautomatingagenticworkflow}.

\subsection{MAS Initialization as Agent Generation}

Based on the previous definition, we formalize the MAS initialization problem as a function \( f_{\text{init}}: q \rightarrow \mathcal{A} \), which maps the user query \( q \) to an initial set of agents \( \mathcal{A} \).
Existing agent generation methods like AutoAgents and EvoAgent are specific implementations of \( f_{\text{init}} \). These methods, which rely solely on LLM interactions for initialization, can be denoted as \( f_{\text{init}}^{\text{LLM}} \). Building on this, AgentInit introduces an additional module that applies multi-objective selection to identify a subset from the results of \( f_{\text{init}}^{\text{LLM}} \) (i.e.\( f_{\text{init}}^{\text{AgentInit}}(q) \subseteq f_{\text{init}}^{\text{LLM}}(q) \)).

\section{AgentInit}
\subsection{Motivation}

Recent studies have shown that MAS often suffers from task derailment and step repetition~\citep{cemri2025multiagentllmsystemsfail}. These issues are often rooted in the initialization phase, where the inclusion of irrelevant or redundant agents can hinder collaboration efficiency and effectiveness.
Drawing on insights from human team collaboration, where task performance is influenced by not only individuals’ ability to handle specific tasks~\citep{devine2001smarter}, but also team diversity to solve complex problems~\citep{nikoleizig2019prediction,higgs2005influence}, we argue that the effectiveness of MAS initialization is governed by two critical dimensions: task relevance and agent diversity.
To this end, we propose \textbf{AgentInit}, a novel initialization framework that constructs effective and efficient agent teams by jointly orchestrating task relevance and agent diversity. This is achieved through two key components: \textbf{Standardized Agent Generation} and \textbf{Balanced Team Selection}, as illustrated in Figure~\ref{fig:MAIN}.

\subsection{Standardized Agent Generation}

Just as onboarding aligns new employees with organizational goals, we standardize agent initialization to enable consistent evaluation and collaboration. To this end, we construct a set of candidate agents \( \mathcal{A}_{\text{candidate}} \) based on the user query \( q \).
Inspired by AutoAgents~\citep{chen2024autoagents}, we introduce a Planner Agent \( G_p \) and an Observer Agent \( G_o \), which interact iteratively to optimize agent generation. Unlike approaches that enforce rigid format constraints, which may limit the Planner's expressiveness and reduce agent quality~\citep{tam2024letspeakfreelystudy}, our method avoids such restrictions during generation.
Instead, a Formatter Agent \( G_f \) is applied after generation to standardize agents, enabling fairer evaluation and more effective selection.

\paragraph{Task Decomposition and Agent Construction}
Planner \( G_p \) serves as the core component in this module. In an MAS, agents typically collaborate by handling different sub-tasks.
Therefore, \( G_p \) is designed to perform task decomposition \( G_{p_1} \) and agent construction \( G_{p_2} \) sequentially, which firstly divides the query into a few sub-tasks, and then generates a capable agent role for each sub-task:
\begin{equation}
G_{p} = (G_{p_2} \circ G_{p_1})
\end{equation}

Specifically, at Round $t$ of the iterative generation process, \( G_{p_1} \) is provided the query $q$ with previous round's feedback $\phi^{t-1}$, as well as the generated sub-tasks set \( \{ \tau_i^{t-1} \}_{i=1}^{n_{t-1}} \) with its corresponding standardized agent set \( \{ \hat{A}_i^{t-1} \}_{i=1}^{n_{t-1}} \), 
where \(n_{t-1}\) denotes the number of sub-tasks determined by $G_{p_1}$ in Round \(t-1\).
Then, $G_{p_1}$ is asked to produce the sub-tasks of the current round:
\begin{multline}
\label{eq:2}
\left\{ \tau_{i}^{t} \right\}_{i=1}^{n_t} = \\
G_{p_1} \left( q, \phi^{t-1}, 
    \left\{ \tau_{i}^{t-1} \right\}_{i=1}^{n_{t-1}}, 
    \left\{ \hat{A}_{i}^{t-1} \right\}_{i=1}^{n_{t-1}} \right)
\end{multline}
Subsequently, \( G_{p_2} \) designs an agent $A_i^t$ for solving each sub-task $\tau_i^t$, expressed in natural language:
\begin{equation}
\left\{ A_{i}^{t} \right\}_{i=1}^{n_t} = G_{p_2} \left( \left\{ \tau_{i}^{t} \right\}_{i=1}^{n_t} \right)
\end{equation}

\paragraph{NL-to-Format Standardization} To standardize the generated agents for format compliance and to facilitate fair evaluation in the subsequent Balanced Team Selection module, we use the Formatter \( G_f \) to convert agents represented in natural language into a standardized format (e.g., JSON):
\begin{equation}
\left \{  \hat{A}_{i}^{t}\right\}_{i=1}^{n_t} = G_{f} \left( \left\{ {A}_{i}^{t} \right\}_{i=1}^{n_t} \right)
\end{equation}
The resulting \( \hat{A}_i^{t} \) denotes the standardized agent representation for sub-task \( \tau_i^{t} \). This conversion ensures both compliance with the required format and improves the quality of the generated agents.

\paragraph{Evaluation and Feedback} At the end of each round, the standardized agents and sub-tasks are reviewed by the Observer Agent \( G_o \). The Observer evaluates the rationality of sub-task decomposition and agent assignment, providing feedback \( \phi^{t} \) as:
\begin{equation}
\phi^{t} = G_{o} \left( \left\{ \hat{A}_{i}^{t} \right\}_{i=1}^{n_t}, \left\{\tau_{i}^{t} \right\}_{i=1}^{n_t}\right)
\end{equation}

\paragraph{Multi-round Iteration} This process is repeated for \( K \) rounds, with the Observer excluded in the final round. The complete Standardized Agent Generation module, denoted as \( G \), is given by:
\begin{equation}
\label{eq:recruitment}
G = (G_{f} \circ G_{p}) \circ \left( (G_{o} \circ G_{f} \circ G_{p})^{K-1} \right)
\end{equation}
The final set of agents \( \left\{ \hat{A}_i^{K} \right\}_{i=1}^{n_K} \) forms the refined candidate set \( \mathcal{A}_{\text{candidate}} \) for the subsequential Balanced Team Selection procedure.

\subsection{Balanced Team Selection}
Expertise of members alongside team diversity plays a vital role in effective problem-solving in human teams.
Therefore, we design the team selection process to balance two key objectives: task relevance and team diversity, aiming to assemble a team that collaborates effectively and achieves strong overall performance.

\paragraph{Candidate Team Construction}
We begin by constructing candidate teams $\mathcal{A'}$ from the full set of generated agents $\mathcal{A}_{\text{candidate}}$, considering all possible combinations whose team size falls within the pre-defined bounds $N_\text{min}$ and $N_\text{max}$:
\begin{equation}
\label{eq:team}
\mathcal{T} = \bigcup_{r = N_{\text{min}}}^{N_{\text{max}}} \left\{ \mathcal{A}' \subseteq \mathcal{A}_{\text{candidate}} : |\mathcal{A}'| = r \right\}
\end{equation}
where we represent the set of all candidate teams as $\mathcal{T}$.
Subsequently, we identify the optimal teams by jointly optimizing two objectives: the team's relevance to the query, \( \text{Rel}(\mathcal{A}', q) \) and the diversity among team members, \( \text{Div}(\mathcal{A}') \).
This problem can be represented as:
\begin{equation} \label{eq:optimize}
\mathcal{T}^* = \left\{ \mathcal{A} \in \mathcal{T} \;\middle|\;
\begin{aligned}
&\nexists\, \mathcal{A}' \in \mathcal{T}, \\
&\text{Rel}(\mathcal{A}', q) \geq \text{Rel}(\mathcal{A}, q) \land \\
&\text{Div}(\mathcal{A}') \geq \text{Div}(\mathcal{A})
\end{aligned}
\right\}
\end{equation}
where \( \mathcal{T^*} \) represents the Pareto optimal set that balances task relevance and diversity, which contains a set of non-dominated agent teams.  A detailed explanation is provided in Appendix~\ref{sec:Pareto}.

\paragraph{Definition of Objectives}
We encode both the agent descriptions and the query into sentence embeddings using a pre-trained text encoder \( \mathcal{E}: \mathcal{L} \rightarrow \mathbb{R}^d \), such as Sentence-BERT \citep{reimers2019sentence} or MiniLM \citep{wang2020minilm}, where $\mathcal{L}$ represents the space of natural language and $d$ is the dimensionality of the resulting embeddings.
Given a candidate team \( \mathcal{A}' \subseteq \mathcal{A}_{\text{candidate}} \) and a query \( q \), the \textbf{relevance} score is defined as the average cosine similarity between the description of each agent member and the query:
\begin{equation}
\label{eq:9}
\text{Rel}(\mathcal{A}', q) = \frac{1}{|\mathcal{A}'|} \sum_{\hat{A} \in \mathcal{A}'} \frac{\mathcal{E}(\hat{A}) \cdot \mathcal{E}(q)}{\|\mathcal{E}(\hat{A})\| \|\mathcal{E}(q)\|}
\end{equation}
where $\hat{A} \in \mathcal{A}'$ is a specific agent member.
To quantify the \textbf{diversity} of a candidate team, we employ the Vendi Score \citep{friedmanvendi}, a metric designed to measure diversity in machine learning applications.
We construct a similarity matrix \( \bm{S}^{|\mathcal{A}'| \times |\mathcal{A}'|} \) for each candidate team $\mathcal{A}'$, whose element \( s_{ij} \) represents the cosine similarity between the embeddings of the description
of \( \hat{A}_i \) and \( \hat{A}_j \):
\begin{equation}
s_{ij} = \frac{\mathcal{E}(\hat{A}_i) \cdot \mathcal{E}(\hat{A}_j)}{\|\mathcal{E}(\hat{A}_i)\| \|\mathcal{E}(\hat{A}_j)\|}
\end{equation}
The Vendi Score is then computed as:
\begin{equation}
\text{Div}(\mathcal{A}') = \exp\left(- \sum_{i=1}^{|\mathcal{A}'|} \lambda_i \log \lambda_i \right)
\end{equation}
where \( \lambda_i \) are the eigenvalues of the similarity matrix \( \bm{S} \) for the candidate team \( \mathcal{A}' \).

\paragraph{Final Team Selection}
After acquiring optimal teams $\mathcal{T^*}$ using Equation (\ref{eq:optimize}), we use a Selector Agent \( G_{s} \) to select the most appropriate agent team for the query:
\begin{equation}
\label{eq:12}
\mathcal{A^*} = G_{s}(\mathcal{T^*}, q)
\end{equation}
With the initialization finalized, the agent set \( \mathcal{A^*} \) is ready to be deployed in different multi-agent inference frameworks, including graph-based interaction structures.
This enables a flexible evaluation of their collaborative performance across a range of scenarios, as discussed in the subsequent experiments.
The algorithmic procedure is provided in Appendix \ref{sec:appendix}, and the prompts used are presented in Appendix \ref{sec:prompt}.

\begin{table*}[htb]
\centering
  \scalebox{0.7}{
\begin{tabular}{l|ccccccc}
\Xhline{1.2pt}
\textbf{Method} & \textbf{MMLU} & \textbf{GSM8K}&\textbf{AQUA} &\textbf{MultiArith} & \textbf{SVAMP} & \textbf{HumanEval} & \textbf{Avg.} \\
\Xhline{1.2pt}
\multicolumn{8}{c}{{Base model: Qwen2.5-72B-Instruct}} \\
\hline
{Vanilla}&81.1 &90.4 &82.1&97.8&92.7 & 84.7&88.1\\
{CoT}&81.4 &92.2  &84.2&\textbf{100.0}&93.4 &84.6 &89.3\\
{AgentPrune}&83.7 & 92.8 &\textbf{85.0}&99.4      &93.2 &\underline{87.6} &\underline{90.3}\\

\(\text{MAS}_\text{none}\)&82.4&92.8&83.4&\textbf{100.0}&93.2&83.5&89.2\\
Pre-defined&82.3 &\underline{93.4} &83.6&\textbf{100.0}&\textbf{93.7} &87.0 &90.0\\
\hdashline[1pt/2pt]
EvoAgent &83.7&\underline{93.4} &84.6&\textbf{100.0}&92.9&83.9&89.8 \\
AutoAgents &\underline{85.3} &92.7 &83.8&\textbf{100.0}&92.9 &86.0& 90.1\\
AgentInit &\textbf{87.3} & \textbf{94.1}&\textbf{85.0}&\textbf{100.0}& \underline{93.5}&\textbf{88.0} &\textbf{91.3}\\

\hline
\multicolumn{8}{c}{{Base model: Deepseek-V3-671B-Instruct}} \\
\hline
{Vanilla}&85.6& 94.5&84.6&\textbf{100.0}& 93.9&88.4&91.2\\
{CoT}&84.3&95.0&85.2&\textbf{100.0}&93.6&89.3&91.2\\
{AgentPrune}&89.5 & 95.3 &86.7&\textbf{100.0}&93.6 &87.2 &92.1\\
 \(\text{MAS}_\text{none}\)&87.6&95.2&86.7&\textbf{100.0}&92.0&87.6&91.5\\
Pre-defined& 88.2 &\underline{95.5}  &87.1&\textbf{100.0}& \textbf{94.6} &  88.5&92.3 \\
\hdashline[1pt/2pt]
EvoAgent &\underline{92.2} &94.9 &\textbf{87.5}&99.4&92.5&88.4&92.5\\
AutoAgents        &90.2 &95.4&86.7&99.4&93.3&\textbf{91.7}&\underline{92.8}\\
AgentInit        &\textbf{92.8} &\textbf{95.7}&\textbf{87.5}&\textbf{100.0}&\underline{94.3}&\textbf{91.7}&\textbf{93.7}\\

\Xhline{1.2pt}
\end{tabular}
}
\caption{Performance comparison between AgentInit and other baseline techniques on Complete Graph.  Best result in each framework is \textbf{bolded}, second-best is \underline{underlined}.}
\label{tab:MAIN}
\end{table*}

\section{Experiments}
\subsection{Experimental Setup}
\label{sec:exp_setup}

\begin{table*}[ht]
  \centering
  \scalebox{0.7}{
\begin{tabular}{l|*{8}{rr}}
\Xhline{1.2pt}
\rowcolor{CadetBlue!20} 

\multirow{2}{*}{\textbf{Method}} 
  & \multicolumn{2}{c}{\textbf{MMLU}} 
  & \multicolumn{2}{c}{\textbf{GSM8K}} 
  & \multicolumn{2}{c}{\textbf{AQuA}} 
  & \multicolumn{2}{c}{\textbf{MultiArith}} 
  & \multicolumn{2}{c}{\textbf{SVAMP}} 
  & \multicolumn{2}{c}{\textbf{HumanEval}} 
  & \multicolumn{2}{c}{\textbf{Avg.}}  \\
\cline{2-15}

 & \textbf{Ptok.} & \textbf{Ctok.} 
  & \textbf{Ptok.} & \textbf{Ctok.} 
  & \textbf{Ptok.} & \textbf{Ctok.} 
  & \textbf{Ptok.} & \textbf{Ctok.} 
  & \textbf{Ptok.} & \textbf{Ctok.} 
  & \textbf{Ptok.} & \textbf{Ctok.} 
  & \textbf{Ptok.} & \textbf{Ctok.}  \\
\Xhline{1.2pt}
Vanilla & 34K& 2K &131K &189K&33K&87K&15K&8K&22K&11K&31K&14K&44K&52K\\
CoT&36K&13K&141K&270K&35K&99K&16K&20K&25K&48K&32K&14K&48K&77K\\
\cdashline{1-15}
AgentPrune & \textcolor[HTML]{8fb1fe}{742K} & \textcolor[HTML]{bbd1f8}{286K} & \textcolor[HTML]{bd1f2d}{7.0M} & \textcolor[HTML]{f7a889}{1.6M} & \textcolor[HTML]{f7b89c}{1.2M} & \textcolor[HTML]{f6a586}{478K} & \textcolor[HTML]{f39778}{805K} & \textcolor[HTML]{80a3fa}{138K} & \textcolor[HTML]{f39778}{1.4M} & \textcolor[HTML]{dbdcde}{246K} & \textcolor[HTML]{485fd1}{257K} & \textcolor[HTML]{465ecf}{51K} & \textcolor[HTML]{d55042}{1.9M} & \textcolor[HTML]{d8dce2}{467K}\\
\(\text{MAS}_\text{none}\) & \textcolor[HTML]{3b4cc0}{240K} & \textcolor[HTML]{3b4cc0}{10K} & \textcolor[HTML]{536edd}{3.7M} & \textcolor[HTML]{3b4cc0}{854K} & \textcolor[HTML]{7a9df8}{780K} & \textcolor[HTML]{3b4cc0}{202K} & \textcolor[HTML]{6180e9}{382K} & \textcolor[HTML]{4961d2}{86K} & \textcolor[HTML]{6e90f2}{669K} & \textcolor[HTML]{6384eb}{150K} & \textcolor[HTML]{b40426}{454K} & \textcolor[HTML]{b40426}{272K} & \textcolor[HTML]{445acc}{1.0M} & \textcolor[HTML]{3b4cc0}{262K}\\
Pre-defined & \textcolor[HTML]{b2ccfb}{938K} & \textcolor[HTML]{c3d5f4}{304K} & \textcolor[HTML]{b40426}{7.1M} & \textcolor[HTML]{e2dad5}{1.4M} & \textcolor[HTML]{f7b89c}{1.2M} & \textcolor[HTML]{d2dbe8}{378K} & \textcolor[HTML]{f4c5ad}{716K} & \textcolor[HTML]{7093f3}{124K} & \textcolor[HTML]{b40426}{1.7M} & \textcolor[HTML]{f3c8b2}{278K} & \textcolor[HTML]{5572df}{266K} & \textcolor[HTML]{3b4cc0}{42K} & \textcolor[HTML]{b40426}{2.0M} & \textcolor[HTML]{b7cff9}{421K}\\
EvoAgent & \textcolor[HTML]{b40426}{2.2M} & \textcolor[HTML]{b40426}{730K} & \textcolor[HTML]{edd2c3}{5.5M} & \textcolor[HTML]{b40426}{1.9M} & \textcolor[HTML]{b40426}{1.5M} & \textcolor[HTML]{b40426}{585K} & \textcolor[HTML]{b40426}{968K} & \textcolor[HTML]{b40426}{384K} & \textcolor[HTML]{e0dbd8}{1.1M} & \textcolor[HTML]{b40426}{381K} & \textcolor[HTML]{3b4cc0}{248K} & \textcolor[HTML]{cbd8ee}{142K} & \textcolor[HTML]{d55042}{1.9M} & \textcolor[HTML]{b40426}{687K}\\
AutoAgents & \textcolor[HTML]{88abfd}{704K} & \textcolor[HTML]{afcafc}{261K} & \textcolor[HTML]{3b4cc0}{3.4M} & \textcolor[HTML]{5470de}{943K} & \textcolor[HTML]{cdd9ec}{1.0M} & \textcolor[HTML]{ccd9ed}{370K} & \textcolor[HTML]{516ddb}{352K} & \textcolor[HTML]{4f69d9}{92K} & \textcolor[HTML]{3b4cc0}{468K} & \textcolor[HTML]{3b4cc0}{115K} & \textcolor[HTML]{6f92f3}{282K} & \textcolor[HTML]{dadce0}{155K} & \textcolor[HTML]{445acc}{1.0M} & \textcolor[HTML]{6788ee}{323K}\\
AgentInit & \textcolor[HTML]{88abfd}{706K} & \textcolor[HTML]{b2ccfb}{267K} & \textcolor[HTML]{3b4cc0}{3.4M} & \textcolor[HTML]{6282ea}{986K} & \textcolor[HTML]{3b4cc0}{603K} & \textcolor[HTML]{7a9df8}{278K} & \textcolor[HTML]{3b4cc0}{300K} & \textcolor[HTML]{3b4cc0}{71K} & \textcolor[HTML]{455cce}{515K} & \textcolor[HTML]{4358cb}{123K} & \textcolor[HTML]{4f69d9}{262K} & \textcolor[HTML]{cad8ef}{141K} & \textcolor[HTML]{3b4cc0}{964K} & \textcolor[HTML]{5e7de7}{311K}\\
\Xhline{1.2pt}
\end{tabular}
}
\caption{Token cost during inference in the Qwen model on Complete Graph. \textbf{Ptok.} indicates the number of prompt tokens used, whereas \textbf{Ctok.} refers to the number of completion tokens produced by the agents. Cooler colors indicate lower token cost, and warmer colors indicate higher cost, within each column.}
\label{tab:main_tokens}
\end{table*} 

\begin{table*}[htb]
\centering
\scalebox{0.65}{
\begin{tabular}{ll|ccccccc}
\Xhline{1.2pt}
\textbf{Framework} & \textbf{Method}& \textbf{MMLU} & \textbf{GSM8K}&\textbf{AQUA} &\textbf{MultiArith} & \textbf{SVAMP} & \textbf{HumanEval} & \textbf{Avg.} \\
\Xhline{1.2pt}
\multirow{5}{*}{Chain} &\(\text{MAS}_\text{none}\)&85.6&94.8&85.4&\textbf{100.0}&\underline{93.6}&88.4&91.3\\
&Pre-defined&90.8&94.9&\underline{87.5}&\textbf{100.0}&\textbf{93.9}&\underline{90.5}&\underline{92.9}\\
&EvoAgent&\underline{91.5}&94.7&\textbf{87.9}&\textbf{100.0}&92.1&86.8&92.2\\
&AutoAgents&91.2&\textbf{95.3}&86.3&\textbf{100.0}&93.2&88.4&92.4\\
&AgentInit&\textbf{92.2}&\underline{95.1}&87.1&\textbf{100.0}&\underline{93.6}&\textbf{90.9}&\textbf{93.2}\\
\hdashline[1pt/2pt]
\multirow{5}{*}{Star} &\(\text{MAS}_\text{none}\)&87.9&94.5&84.6&\textbf{100.0}&92.7&89.2&91.5\\
&Pre-defined&89.5&\underline{95.2}&85.0&\textbf{100.0}&\textbf{93.5}&\underline{90.1}&\underline{92.2}\\
&EvoAgent&\textbf{91.9}&94.4&\textbf{85.4}&\textbf{100.0}&91.4&87.6&91.8\\
&AutoAgents&89.5&95.1&\textbf{85.4}&\textbf{100.0}&91.8&{90.1}&92.0\\
&AgentInit&\textbf{91.9}&\textbf{95.9}&\textbf{85.4}&\textbf{100.0}&\underline{93.2}&\textbf{91.7}&\textbf{93.0}\\
\hdashline[1pt/2pt]
\multirow{5}{*}{Layered} &\(\text{MAS}_\text{none}\)&86.9&\underline{95.5}&87.9&98.8&\underline{92.9}&86.8&91.5\\
&Pre-defined&88.9&\underline{95.5}&85.8&\textbf{100.0}&\textbf{93.6}&\underline{89.3}&92.2\\
&EvoAgent&90.8&95.3&\underline{88.3}&\textbf{100.0}&\underline{92.9}&87.6&92.5\\
&AutoAgents&\textbf{93.5}&\underline{95.5}&87.1&\textbf{100.0}&92.5&88.4&\underline{92.8}\\
&AgentInit&\textbf{93.5}&\textbf{95.7}&\textbf{89.2}&\textbf{100.0}&\underline{92.9}&\textbf{91.7}&\textbf{93.8}\\
\hdashline[1pt/2pt]
\multirow{5}{*}{AutoGen} &\(\text{MAS}_\text{none}\)&88.3&95.2&86.3&98.8&91.8&85.1&90.9\\
&Pre-defined&91.7 &94.9&\underline{87.1}&99.4&\textbf{93.6}&88.4&92.5  \\
&EvoAgent &90.8&94.0&\underline{87.1}&\textbf{100.0}&93.2&\underline{90.1}&92.5\\
&AutoAgents        &\underline{92.0}&\textbf{95.8}&86.7&\textbf{100.0}&92.5&88.4&\underline{92.6}\\
&AgentInit        &\textbf{92.2}&\underline{95.4}&\textbf{88.3}&\textbf{100.0}&\textbf{93.6}&\textbf{93.0}&\textbf{93.8}\\

\Xhline{1.2pt}
\end{tabular}}
\caption{Performance comparison between AgentInit and other MAS initialization methods across different MAS frameworks, evaluated on Deepseek-V3. Best result in each framework is \textbf{bolded}, second-best is \underline{underlined}.}
\label{tab:main_differentF}
\end{table*}

\paragraph{Models and Benchmarks}
We evaluate AgentInit on Qwen2.5-72B-Instruct~\citep{qwen2.5} and Deepseek-V3-671B-Instruct-1226~\citep{liu2024deepseek}. General reasoning is assessed with MMLU~\citep{hendrycks2021measuring}; math reasoning with GSM8k~\citep{cobbe2021gsm8k}, MultiArith~\citep{roy-roth-2015-solving}, AQuA~\citep{patel-etal-2021-nlp}, and SVAMP~\citep{ling2017program}; and code generation with HumanEval~\citep{chen2021codex}. 
For complex tasks, we use MATH~\citep{hendrycksmath2021} and AIME2025 for advanced mathematics problems, as well as Trivia Creative Writing~\citep{wang2024unleashingemergentcognitivesynergy} and ScienceWorld~\citep{wang2022scienceworld}, which require knowledge-grounded storytelling and interactive scientific reasoning. Detailed descriptions and results of all complex datasets are provided in Appendix~\ref{sec:science}.

\paragraph{MAS Frameworks}  
We explored different MAS frameworks, using AutoGen for Loosely Modeled Frameworks and graph structures for Formally Modeled Frameworks. Specifically, we adopted the graph framework from Agentprune~\citep{zhang2025cut}, including chain, star, layered, and complete graph structures.

\paragraph{Baselines}
For tasks performed by a single agent, we compare our method with the direct reasoning (referred to as Vanilla) and the Chain-of-Thought approach (CoT,~\citealp{cot}). We also include the full {Agentprune} method as an MAS baseline. Moreover, we compare several MAS initialization strategies across different frameworks, including a strategy without any role assignment (denoted as $\text{MAS}_\text{none}$) as well as automatic initialization methods of AutoAgents, EvoAgent, and the pre-defined agents used in {Agentprune}.

\paragraph{Implementation Details}
For experiments with Qwen2.5 and Deepseek-V3, we perform model inference via API with a temperature of 1, following AgentDropout~\citep{wang2025agentdropoutdynamicagentelimination}. We use one inference round for the graph structure. The number of iteration rounds is set to \(K = 3\), which is adopted in Equation~(\ref{eq:recruitment}). The number of generated agents is set between \(N_{\text{min}} = 1\) and \(N_{\text{max}} = 5\) in Equation~(\ref{eq:team}). The text encoder \(\mathcal{E}\) is implemented using all-MiniLM-L6-v2~\citep{wang2020minilm}.

\subsection{Main Result}
\paragraph{AgentInit demonstrates superior performance and efficiency across benchmarks.}
As shown in Table~\ref{tab:MAIN}, AgentInit consistently outperforms existing approaches, including CoT, state-of-the-art (SOTA)  MAS initialization methods like AutoAgents and EvoAgent, and pre-defined strategies. 
Although pre-defined MAS occasionally show better synergy on related datasets, AgentInit still outperforms SOTA methods by 1.2 and 0.9 points, and pre-defined approaches by 1.3 and 1.4 points on Qwen2.5 and Deepseek-V3, respectively, in overall performance.
AgentInit also achieves the best results on ScienceWorld and Trivia Creative Writing tasks (Appendix~\ref{sec:science}).
Moreover, Table~\ref{tab:main_tokens} shows that AgentInit reduces prompt token and completion token usage during inference. Furthermore, we conducted additional experiments with larger maximum team sizes. The results show that AgentInit’s efficiency advantage becomes more pronounced as $N_{max}$ increases, confirming the scalability of our approach (Appendix~\ref{sec:efficiency}). This efficiency comes from Balanced Team Selection, which filters out redundant agents and retains only those contributing to task performance. 
These results highlight AgentInit's effectiveness and efficiency across diverse settings. Case studies showcasing the generated agents and their reasoning results are provided in the Appendix ~\ref{sec:case}.

\paragraph{AgentInit is robust to MAS frameworks.}
AgentInit demonstrates strong adaptability across a variety of MAS frameworks, including both graph-based and AutoGen-style architectures, and consistently achieves superior overall performance, as evidenced by Table \ref{tab:MAIN} and Table \ref{tab:main_differentF}.
Notably, within the Complete Graph framework--which contains the most comprehensive information and thus offers the highest performance ceiling--AgentInit further refines MAS team composition effectively.
By minimizing the influence of redundant or detrimental agents, it enables the framework to more fully exploit its collaborative potential.
For more details on Qwen's performance across different frameworks, please refer to Appendix~\ref{sec:qwen}.

\section{Analysis}



\begin{table*}[htb]
\centering
  \scalebox{0.7}{
\begin{tabular}{lccccccc}
\Xhline{1.2pt}
  \textbf{Setup}
  & \textbf{MMLU} 
  & \textbf{GSM8K} 
  & \textbf{AQuA} 
  & \textbf{MultiArith} 
  & \textbf{SVAMP} 
  & \textbf{HumanEval} 
  & \textbf{Avg.}  \\
\Xhline{1.2pt}
AgentInit &\textbf{87.3} & 94.1&\textbf{85.0}&\textbf{100.0}& 93.5&88.0 & \textbf{91.3}\\
\hline

\multicolumn{8}{c}{\textbf{Iteration Rounds (Default: \(K\) = 3)}}\\
\hdashline
\(K\) = 1&85.6&94.2&83.8&100.0&93.2&87.6&90.7 \\
\(K \) = 5&87.3&94.0&84.4&100.0&93.6&88.0&91.2 \\
\hline
\multicolumn{8}{c}{\textbf{Agent Standardization Method (Default: NL-to-Format)}}\\
\hdashline
None&{87.6}&93.7&84.2&{100.0}&91.8&88.4&91.0\\
\hline
\multicolumn{8}{c}{\textbf{Selection Objectives (Default: Rel, Div)}}\\
\hdashline
Only Rel&85.7&94.0&84.0&{100.0}&92.5&86.3&90.4\\
Only Div&86.3&94.0&83.5&{100.0}&92.9&86.0&90.5\\
Rel, Div\(_\text{avg}\)&86.0&{94.1}&{85.0}&{100.0}&93.2&87.6&91.0\\
\hline 
\multicolumn{8}{c}{\textbf{Selection Strategy (Default: Pareto Best)}}\\
\hdashline
None &85.6&93.4&84.2&100.0&92.7&85.6&90.2\\
Global Worst &83.7&92.6 &83.3&100.0&92.5&81.8&89.0\\
Pareto Worst &84.4&92.4&83.3&100.0&92.1&85.1&89.6\\
Random &84.4&94.1&83.8&100.0&92.9&86.4&90.3\\
\Xhline{1.2pt}
\end{tabular}
}
\caption{Comparison results of different setups.}
\label{tab:analyze_ablation}
\end{table*}

\begin{table*}[htb]
\centering
\scalebox{0.7}{
\begin{tabular}{lccccccc}
\Xhline{1.2pt}
  \textbf{Method}
  & \textbf{MMLU} 
  & \textbf{GSM8K} 
  & \textbf{AQuA} 
  & \textbf{MultiArith} 
  & \textbf{SVAMP} 
  & \textbf{HumanEval} 
  & \textbf{Avg.}  \\
\Xhline{1.2pt}
\(\text{MAS}_\text{none}\)&82.4&92.8&83.4&\textbf{100.0}&93.2&83.5&89.2\\
\(\text{AgentInit}_{(1,1)}\) &83.4&93.9&84.1&\textbf{100.0}&92.7&86.8&90.2\\
\(\text{AgentInit}_{(1,10)}\) &85.0&94.0&84.6&\textbf{100.0}&93.3&83.5&90.1\\
\(\text{AgentInit}_{(10,10)}\) & 83.6&\textbf{94.1}&84.8&\textbf{100.0}&93.3&85.1&90.2\\
AgentInit        &\textbf{87.3} & \textbf{94.1}&\textbf{85.0}&\textbf{100.0}& \textbf{93.5}&\textbf{88.0} & \textbf{91.3}\\
\Xhline{1.2pt}

\end{tabular}
}
\caption{Transferability of AgentInit with single query initialization.}
\label{tab:analyze_transfer}
\end{table*}
\begin{table*}[t]
\centering
\scalebox{0.7}{
\begin{tabular}{lrrrr}
\toprule
\textbf{Dataset} & \textbf{Total GPU Time (s)} & \textbf{Avg per Sample (s)} & \textbf{Iterations (Avg)} & \textbf{Team Size (Avg)} \\
\midrule
MMLU       & 16.2   & 0.0162 & 2.19 & 2.69 \\
HumanEval  & 10.7   & 0.0106 & 2.25 & 2.28 \\
GSM8K      & 13.7   & 0.0107 & 2.29 & 2.66 \\
MultiArith & 13.0   & 0.0130 & 2.08 & 2.18 \\
\bottomrule
\end{tabular}
}
\caption{GPU time, average number of iterations, and average team size for the team selection process with $N_{max}=5$.}
\label{tab:team-selection-efficiency}
\end{table*}

In this section, we conduct experiments using the Qwen2.5 model, following the same settings outlined in Section~\ref{sec:exp_setup}, with the framework instantiated under the Complete Graph configuration.

\subsection{Ablation Study}
\label{sec:ablation}
\paragraph{Effect of iteration rounds}
To investigate the impact of the iteration count \(K\) during Standardized Agent Generation on MAS initialization quality, we conducted experiments across varying \(K\) values (see the \textit{\textbf{Iteration Rounds}} block in Table~\ref{tab:analyze_ablation})\footnote{When \(K=1\), the Observer Agent \(G_o\) is excluded from the generation process.}.
The results indicate that the performance at \(K=3\) and \(K=5\) is nearly identical, differing by only 0.1\%.
As the best performance is achieved at \(K=3\), we adopt this setting in all main experiments.
These findings suggest that iterative refinement, facilitated by incorporating feedback through $G_o$, enhances initialization quality to a certain extent, but additional interactions offer diminishing returns.
This outcome also reflects the inherent limitations of generating agents exclusively through LLM-based interactions.

\paragraph{Effect of Agent Standardization}  
To assess the impact of NL-to-Format standardization, we conduct ablation experiments on the Formatter \(G_f\).
As shown in the \textit{\textbf{Agent Standardization Method}} block of Table~\ref{tab:analyze_ablation}, omitting this procedure leads to a 0.4-point drop in overall performance.
This decline highlights the significance of this standardization step in ensuring consistency and fairness while comparing different candidate teams.

\paragraph{Effect of Selection Objectives}  
To assess the impact of selection objectives in Balanced Team Selection, we conduct ablation studies using single-objective strategies based on relevance and diversity, denoted as \textit{Only Rel} and \textit{Only Div} in the \textit{\textbf{Selection Objectives}} block of Table~\ref{tab:analyze_ablation}. We also evaluate an alternative diversity metric, \textit{\(\text{Div}_\text{avg}\)}, which minimizes the average pairwise similarity in the similarity matrix \(\bm{S}\).
Results show that \textit{\(\text{Div}_\text{avg}\)} improves performance by 0.6 points over the relevance-only setting, but still underperforms compared to the full AgentInit. These findings confirm the effectiveness of the proposed relevance and diversity objectives for team selection.

\paragraph{Effect of Selection Strategy}  
To verify the effectiveness of each component in the Balanced Team Selection module, we conduct a series of controlled experiments, with the results shown in the \textit{\textbf{Selection Strategy}} block of Table \ref{tab:analyze_ablation}.
\textit{None} indicates that no agent selection is performed and all candidate agents are used directly.
This setting results in a 1.1-point drop in performance compared to the default configuration, demonstrating the importance of the selection process in enhancing overall effectiveness.
\textit{Pareto Worst} denotes intentionally choosing the worst-performing agent combination within the Pareto optimal set, leading to a substantial performance degradation, further validating the necessity of the selector component.
Moreover, its performance surpasses \textit{Global Worst}, suggesting that generating a Pareto optimal set already raises the lower bound of performance.
\textit{Random} corresponds to randomly selecting the same number of agents used in the AgentInit setting.
Its inferior performance 
indicates that the observed gains cannot be attributed merely to reducing team size.
Instead, they result from our selection strategy itself, which is both reasonable and targeted.

\subsection{Transferability}
Initializing a dedicated agent set for each query incurs significant computational overhead. 
To alleviate this, we evaluate strategies that construct the MAS from a single query or a small batch of queries. 
We denote these strategies as \textbf{AgentInit$(x,y)$}, where $x$ is the number of queries used in the Standardized Agent Generation module, in Equation~(\ref{eq:2}), and $y$ is the number of queries used in the Balanced Team Selection module,in Equation~(\ref{eq:9}) and Equation~(\ref{eq:12}).

As shown in Table~\ref{tab:analyze_transfer}, constructing the MAS using either a single representative query or a small batch of queries yields performance comparable to more complex per-query initialization. 
Although slightly lower than full per-query initialization, these approaches still provide substantial improvements over $MAS_{none}$. 
These results highlight the strong transferability of AgentInit, demonstrating its ability to generalize across similar tasks while supporting flexible initialization strategies.

\subsection{Scalability and Efficiency of Team Selection}
Enumerating all possible agent combinations in team selection may seem expensive, but it is lightweight when the number of candidate agents is small. 
In our experiments with $N_{max}=5$, computing all role combinations and their objective scores is efficient. 
For example, using Qwen2.5-72B-Instruct and extracting role embeddings with all-MiniLM-L6-v2 (on a single NVIDIA 3090 GPU), the average GPU time per sample is negligible. 
Detailed efficiency measurements and team sizes are reported in Table~\ref{tab:team-selection-efficiency}, confirming that the selection process incurs minimal overhead compared to agent generation and inference.

To evaluate scalability to larger candidate sets, experiments were conducted with $N_{max}=10$. Even in this setting, enumerating all combinations took only 8.4738 seconds per sample, which is modest relative to the entire multi-agent inference pipeline. For larger numbers of agents, heuristic approaches such as NSGA-II~\citep{nsgaii} can efficiently approximate the Pareto-optimal set. NSGA-II has an overall time complexity of $O(G N_p^2)$, where $G$ is the number of generations and $N_p$ is the population size. In experiments with a population size of 100 and 50 generations, NSGA-II required only 0.9806 seconds per sample while achieving 80\% coverage of the global Pareto frontier. The Generational Distance (GD) was 0.02, indicating a high-quality approximation.

\subsection{Effects of Objectives on Performance}
\begin{figure}[t]
    \centering
    \includegraphics[width=1\linewidth, trim=9 16 0 0, clip]{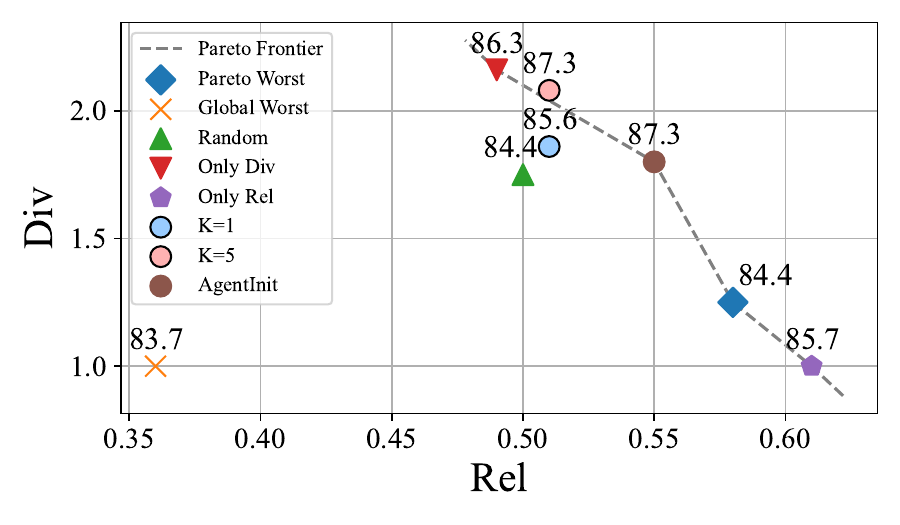} 
    \caption{The impact of diversity and relevance on performance under different settings on the MMLU dataset.}
    \label{fig:analysis}
\end{figure}

To further investigate the impact of diversity and relevance on performance, we measured both metrics under different settings described in Section~\ref{sec:ablation} using the MMLU dataset.
For consistency, we first focused on configurations with \(K=3\) and plotted the Pareto frontier for these settings, as shown in Figure~\ref{fig:analysis}.
The results indicate that performance does not monotonically increase with either diversity or relevance alone, suggesting that optimizing for a single objective may not reliably improve overall performance.

From the Pareto frontier score distribution, it is observed that the highest scores are concentrated in the middle region, where diversity and relevance are balanced. In contrast, extreme optimization towards a single objective corresponds to poorer performance. Settings outside the Pareto optimal set also perform worse, supporting the effectiveness of our selection strategy.

Building on this analysis, we examined how the number of iterations \(K\) affects both objectives and overall performance. Compared to \(K=1\), where no observer is involved, the default setting shows a significant improvement in relevance, indicating that the observer optimizes agent quality. As \(K\) increases to \(K=5\), more agents are generated, trading some relevance for diversity, yet \(K=5\) still outperforms \(K=1\) on both objectives, yielding better overall performance.
\begin{table*}[t]
\centering
\scalebox{0.8}{
\begin{tabular}{lccccc}
\toprule
 & \textbf{MMLU} & \textbf{GSM8K} & \textbf{MultiArith} &\textbf{HumanEval} & \textbf{Avg.} \\
\midrule
Unoptimized Max Similarity (\%) & 67.52 & 56.98 & 50.45 & 59.99 & 58.73 \\
Optimized Max Similarity (\%)   & 65.00 & 53.21 & 45.13 & 56.49 & 54.96 \\
\bottomrule
\end{tabular}
}
\caption{Maximum pairwise similarity (Max($s_{ij}$)) within teams before and after Balanced Team Selection, showing reduced redundancy across datasets.}
\label{tab:team_redundancy}
\end{table*}

\subsection{Effect of Team Redundancy Reduction}  

To further evaluate the impact of the selection objectives on team composition, we measured the maximum pairwise similarity within each team (with diagonal set to 0), denoted as Max($s_{ij}$), where $s_{ij}$ is the cosine similarity between agent embeddings. This metric reflects potential redundancy among agents. Experiments were conducted on representative datasets from knowledge-intensive, mathematical, and programming domains, including MMLU, GSM8K, MultiArith, and HumanEval.

As shown in Table~\ref{tab:team_redundancy}, the maximum pairwise similarity within teams decreases after optimization, with an average reduction of 3.5\% across datasets. This demonstrates that the Balanced Team Selection strategy effectively reduces redundancy, yielding more complementary agent teams and supporting improved overall performance.

\section{Related Work}
\subsection{LLM-based MAS}
LLM-based MAS have emerged as a promising framework for solving complex tasks across domains. Early studies show the potential of LLM-driven agents in collaborative problem-solving~\citep{hao2023chatllmnetworkbrainsintelligence, du2024improving, liang2024encouraging}, paving the way for research in areas such as collaborative programming~\citep{ishibashi2024selforganizedagentsllmmultiagent, hong2024metagpt, qian-etal-2024-chatdev}, embodied teamwork in virtual or physical environments~\citep{guo2024embodied}, and cooperative reasoning through strategic role allocation~\citep{li2023camel, chen2023agentversefacilitatingmultiagentcollaboration, wang2024unleashingemergentcognitivesynergy}. However, manually defined roles still constrain agent adaptability in dynamic settings.
We propose AgentInit, an autonomous MAS initialization method that lets agents configure roles and behaviors based on task and environment.

\subsection{MAS Initialization}
Prior work such as DsPy~\citep{khattab2024dspy} and EvoPrompt~\citep{guo2024connecting} has explored automating prompt generation. 
Further efforts have aimed at automating agent generation to initialize MAS and overcome the limitations of manual configurations in collaborative tasks~\citep{deshpande2023toxicity, xu2025expertpromptinginstructinglargelanguage}. Methods like AgentVerse \citep{chen2023agentversefacilitatingmultiagentcollaboration}, AutoAgents \citep{chen2024autoagents}, and EvoAgent \citep{yuan2025evoagentautomaticmultiagentgeneration} aim to address this by automating agent generation. However, in existing approaches, agents are often generated directly through LLM interactions without further validation, which may hinder MAS performance \citep{wang2025agentdropoutdynamicagentelimination}. While recent methodologies like AFlow \citep{zhang2025aflowautomatingagenticworkflow} and EvoFlow \citep{zhang2025evoflowevolvingdiverseagentic} automate the entire agent workflow, they still rely on additional training data. In contrast, AgentInit adopts an approach that first generates standardized agents and then applies team selection to enhance efficiency without requiring extra data.

\section{Conclusion}
Drawing inspiration from real-world strategies for optimizing team structures, we introduce AgentInit, a novel approach for MAS initialization with strong transferability across similar tasks. By jointly optimizing at both the individual and team levels, AgentInit ensures that agents are highly aligned with task objectives and that the resulting teams exhibit clear division of labor and effective collaboration. Experimental results across a diverse set of tasks and MAS frameworks show that AgentInit consistently enhances system performance and token consumption. These findings provide practical insights for future work on MAS initialization.

\section*{Limitations}
The experiments conducted so far provide preliminary evidence supporting the potential of AgentInit. However, the scope of these experiments remains limited, and further validation is needed to assess its performance across a wider range of tasks and more complex domains. While the evaluation objectives used provide an initial indication of team coordination and overall system effectiveness, their ability to capture subtle aspects of collaboration and performance requires further refinement. Moreover, the automated initialization process of the MAS, though effective in reducing manual intervention, results in significant token overhead and is heavily reliant on the capabilities of high-performing language models. This dependency may present challenges in resource-constrained environments where computational costs are crucial considerations.

\section*{Acknowledgments}
This work was supported in part by Guangdong S\&T Program (Grant No. 2024B0101050003), Guangdong Basic and Applied Basic Research Foundation (Grant No. 2024A1515011491), and Shenzhen Science and Technology Program (Grant Nos. ZDSYS20230626091203008, KJZD20231023094700001, KQTD20240729102154066). 
We would like to thank the anonymous reviewers and meta-reviewer for their insightful suggestions.

\bibliography{custom}

\begin{thebibliography}{60}
\providecommand{\natexlab}[1]{#1}

\bibitem[{Chen et~al.(2024{\natexlab{a}})Chen, Dong, Shu, Zhang, Sesay, Karlsson, Fu, and Shi}]{chen2024autoagents}
Guangyao Chen, Siwei Dong, Yu~Shu, Ge~Zhang, Jaward Sesay, B{\"{o}}rje Karlsson, Jie Fu, and Yemin Shi. 2024{\natexlab{a}}.
\newblock \href {https://www.ijcai.org/proceedings/2024/3} {Autoagents: {A} framework for automatic agent generation}.
\newblock In \emph{Proceedings of the Thirty-Third International Joint Conference on Artificial Intelligence, {IJCAI} 2024, Jeju, South Korea, August 3-9, 2024}, pages 22--30. ijcai.org.

\bibitem[{Chen et~al.(2021)Chen, Tworek, Jun, Yuan, de~Oliveira~Pinto, Kaplan, Edwards, Burda, Joseph, Brockman, Ray, Puri, Krueger, Petrov, Khlaaf, Sastry, Mishkin, Chan, Gray, Ryder, Pavlov, Power, Kaiser, Bavarian, Winter, Tillet, Such, Cummings, Plappert, Chantzis, Barnes, Herbert-Voss, Guss, Nichol, Paino, Tezak, Tang, Babuschkin, Balaji, Jain, Saunders, Hesse, Carr, Leike, Achiam, Misra, Morikawa, Radford, Knight, Brundage, Murati, Mayer, Welinder, McGrew, Amodei, McCandlish, Sutskever, and Zaremba}]{chen2021codex}
Mark Chen, Jerry Tworek, Heewoo Jun, Qiming Yuan, Henrique~Ponde de~Oliveira~Pinto, Jared Kaplan, Harri Edwards, Yuri Burda, Nicholas Joseph, Greg Brockman, Alex Ray, Raul Puri, Gretchen Krueger, Michael Petrov, Heidy Khlaaf, Girish Sastry, Pamela Mishkin, Brooke Chan, Scott Gray, and 39 others. 2021.
\newblock \href {https://arxiv.org/abs/2107.03374} {Evaluating large language models trained on code}.
\newblock \emph{ArXiv preprint}, abs/2107.03374.

\bibitem[{Chen et~al.(2024{\natexlab{b}})Chen, Su, Zuo, Yang, Yuan, Chan, Yu, Lu, Hung, Qian, Qin, Cong, Xie, Liu, Sun, and Zhou}]{chen2023agentversefacilitatingmultiagentcollaboration}
Weize Chen, Yusheng Su, Jingwei Zuo, Cheng Yang, Chenfei Yuan, Chi{-}Min Chan, Heyang Yu, Yaxi Lu, Yi{-}Hsin Hung, Chen Qian, Yujia Qin, Xin Cong, Ruobing Xie, Zhiyuan Liu, Maosong Sun, and Jie Zhou. 2024{\natexlab{b}}.
\newblock \href {https://openreview.net/forum?id=EHg5GDnyq1} {Agentverse: Facilitating multi-agent collaboration and exploring emergent behaviors}.
\newblock In \emph{The Twelfth International Conference on Learning Representations, {ICLR} 2024, Vienna, Austria, May 7-11, 2024}. OpenReview.net.

\bibitem[{Cobbe et~al.(2021)Cobbe, Kosaraju, Bavarian, Chen, Jun, Kaiser, Plappert, Tworek, Hilton, Nakano, Hesse, and Schulman}]{cobbe2021gsm8k}
Karl Cobbe, Vineet Kosaraju, Mohammad Bavarian, Mark Chen, Heewoo Jun, Lukasz Kaiser, Matthias Plappert, Jerry Tworek, Jacob Hilton, Reiichiro Nakano, Christopher Hesse, and John Schulman. 2021.
\newblock \href {https://arxiv.org/abs/2110.14168} {Training verifiers to solve math word problems}.
\newblock \emph{ArXiv preprint}, abs/2110.14168.

\bibitem[{Coello(2006)}]{coello2006evolutionary}
C.~A. Coello. 2006.
\newblock \href {https://doi.org/10.1109/MCI.2006.1597059} {Evolutionary multi-objective optimization: a historical view of the field}.
\newblock \emph{Comp. Intell. Mag.}, 1(1):28–36.

\bibitem[{Deb(2011)}]{deb2011multi}
Kalyanmoy Deb. 2011.
\newblock \href {https://doi.org/10.1007/978-0-85729-652-8_1} {\emph{Multi-objective Optimisation Using Evolutionary Algorithms: An Introduction}}, pages 3--34.
\newblock Springer London, London.

\bibitem[{Deb et~al.(2002)Deb, Pratap, Agarwal, and Meyarivan}]{nsgaii}
Kalyanmoy Deb, Amrit Pratap, Sameer Agarwal, and TAMT Meyarivan. 2002.
\newblock A fast and elitist multiobjective genetic algorithm: Nsga-ii.
\newblock \emph{IEEE transactions on evolutionary computation}, 6(2):182--197.

\bibitem[{Deshpande et~al.(2023)Deshpande, Murahari, Rajpurohit, Kalyan, and Narasimhan}]{deshpande2023toxicity}
Ameet Deshpande, Vishvak Murahari, Tanmay Rajpurohit, Ashwin Kalyan, and Karthik Narasimhan. 2023.
\newblock \href {https://doi.org/10.18653/v1/2023.findings-emnlp.88} {Toxicity in chatgpt: Analyzing persona-assigned language models}.
\newblock In \emph{Findings of the Association for Computational Linguistics: EMNLP 2023}, pages 1236--1270, Singapore. Association for Computational Linguistics.

\bibitem[{Devine and Philips(2001)}]{devine2001smarter}
Dennis~J Devine and Jennifer~L Philips. 2001.
\newblock \href {https://doi.org/10.1177/104649640103200501} {Do smarter teams do better: A meta-analysis of cognitive ability and team performance}.
\newblock \emph{Small group research}, 32(5):507--532.

\bibitem[{Du et~al.(2024)Du, Li, Torralba, Tenenbaum, and Mordatch}]{du2024improving}
Yilun Du, Shuang Li, Antonio Torralba, Joshua~B. Tenenbaum, and Igor Mordatch. 2024.
\newblock \href {https://openreview.net/forum?id=zj7YuTE4t8} {Improving factuality and reasoning in language models through multiagent debate}.
\newblock In \emph{Forty-first International Conference on Machine Learning, {ICML} 2024, Vienna, Austria, July 21-27, 2024}. OpenReview.net.

\bibitem[{Friedman and Dieng(2023)}]{friedmanvendi}
Dan Friedman and Adji~Bousso Dieng. 2023.
\newblock \href {https://openreview.net/forum?id=g97OHbQyk1} {The vendi score: A diversity evaluation metric for machine learning}.
\newblock \emph{Transactions on Machine Learning Research}.

\bibitem[{Grattafiori et~al.(2024)Grattafiori, Dubey, Jauhri, Pandey, Kadian, Al-Dahle, Letman, Mathur, Schelten, Vaughan, Yang, Fan, Goyal, Hartshorn, Yang, Mitra, Sravankumar, Korenev, Hinsvark, Rao, Zhang, Rodriguez, Gregerson, Spataru, Roziere, Biron, Tang, Chern, Caucheteux, Nayak, Bi, Marra, McConnell, Keller, Touret, Wu, Wong, Ferrer, Nikolaidis, Allonsius, Song, Pintz, Livshits, Wyatt, Esiobu, Choudhary, Mahajan, Garcia-Olano, Perino, Hupkes, Lakomkin, AlBadawy, Lobanova, Dinan, Smith, Radenovic, Guzmán, Zhang, Synnaeve, Lee, Anderson, Thattai, Nail, Mialon, Pang, Cucurell, Nguyen, Korevaar, Xu, Touvron, Zarov, Ibarra, Kloumann, Misra, Evtimov, Zhang, Copet, Lee, Geffert, Vranes, Park, Mahadeokar, Shah, van~der Linde, Billock, Hong, Lee, Fu, Chi, Huang, Liu, Wang, Yu, Bitton, Spisak, Park, Rocca, Johnstun, Saxe, Jia, Alwala, Prasad, Upasani, Plawiak, Li, Heafield, Stone, El-Arini, Iyer, Malik, Chiu, Bhalla, Lakhotia, Rantala-Yeary, van~der Maaten, Chen, Tan, Jenkins, Martin, Madaan, Malo, Blecher,
  Landzaat, de~Oliveira, Muzzi, Pasupuleti, Singh, Paluri, Kardas, Tsimpoukelli, Oldham, Rita, Pavlova, Kambadur, Lewis, Si, Singh, Hassan, Goyal, Torabi, Bashlykov, Bogoychev, Chatterji, Zhang, Duchenne, Çelebi, Alrassy, Zhang, Li, Vasic, Weng, Bhargava, Dubal, Krishnan, Koura, Xu, He, Dong, Srinivasan, Ganapathy, Calderer, Cabral, Stojnic, Raileanu, Maheswari, Girdhar, Patel, Sauvestre, Polidoro, Sumbaly, Taylor, Silva, Hou, Wang, Hosseini, Chennabasappa, Singh, Bell, Kim, Edunov, Nie, Narang, Raparthy, Shen, Wan, Bhosale, Zhang, Vandenhende, Batra, Whitman, Sootla, Collot, Gururangan, Borodinsky, Herman, Fowler, Sheasha, Georgiou, Scialom, Speckbacher, Mihaylov, Xiao, Karn, Goswami, Gupta, Ramanathan, Kerkez, Gonguet, Do, Vogeti, Albiero, Petrovic, Chu, Xiong, Fu, Meers, Martinet, Wang, Wang, Tan, Xia, Xie, Jia, Wang, Goldschlag, Gaur, Babaei, Wen, Song, Zhang, Li, Mao, Coudert, Yan, Chen, Papakipos, Singh, Srivastava, Jain, Kelsey, Shajnfeld, Gangidi, Victoria, Goldstand, Menon, Sharma, Boesenberg,
  Baevski, Feinstein, Kallet, Sangani, Teo, Yunus, Lupu, Alvarado, Caples, Gu, Ho, Poulton, Ryan, Ramchandani, Dong, Franco, Goyal, Saraf, Chowdhury, Gabriel, Bharambe, Eisenman, Yazdan, James, Maurer, Leonhardi, Huang, Loyd, Paola, Paranjape, Liu, Wu, Ni, Hancock, Wasti, Spence, Stojkovic, Gamido, Montalvo, Parker, Burton, Mejia, Liu, Wang, Kim, Zhou, Hu, Chu, Cai, Tindal, Feichtenhofer, Gao, Civin, Beaty, Kreymer, Li, Adkins, Xu, Testuggine, David, Parikh, Liskovich, Foss, Wang, Le, Holland, Dowling, Jamil, Montgomery, Presani, Hahn, Wood, Le, Brinkman, Arcaute, Dunbar, Smothers, Sun, Kreuk, Tian, Kokkinos, Ozgenel, Caggioni, Kanayet, Seide, Florez, Schwarz, Badeer, Swee, Halpern, Herman, Sizov, Guangyi, Zhang, Lakshminarayanan, Inan, Shojanazeri, Zou, Wang, Zha, Habeeb, Rudolph, Suk, Aspegren, Goldman, Zhan, Damlaj, Molybog, Tufanov, Leontiadis, Veliche, Gat, Weissman, Geboski, Kohli, Lam, Asher, Gaya, Marcus, Tang, Chan, Zhen, Reizenstein, Teboul, Zhong, Jin, Yang, Cummings, Carvill, Shepard, McPhie,
  Torres, Ginsburg, Wang, Wu, U, Saxena, Khandelwal, Zand, Matosich, Veeraraghavan, Michelena, Li, Jagadeesh, Huang, Chawla, Huang, Chen, Garg, A, Silva, Bell, Zhang, Guo, Yu, Moshkovich, Wehrstedt, Khabsa, Avalani, Bhatt, Mankus, Hasson, Lennie, Reso, Groshev, Naumov, Lathi, Keneally, Liu, Seltzer, Valko, Restrepo, Patel, Vyatskov, Samvelyan, Clark, Macey, Wang, Hermoso, Metanat, Rastegari, Bansal, Santhanam, Parks, White, Bawa, Singhal, Egebo, Usunier, Mehta, Laptev, Dong, Cheng, Chernoguz, Hart, Salpekar, Kalinli, Kent, Parekh, Saab, Balaji, Rittner, Bontrager, Roux, Dollar, Zvyagina, Ratanchandani, Yuvraj, Liang, Alao, Rodriguez, Ayub, Murthy, Nayani, Mitra, Parthasarathy, Li, Hogan, Battey, Wang, Howes, Rinott, Mehta, Siby, Bondu, Datta, Chugh, Hunt, Dhillon, Sidorov, Pan, Mahajan, Verma, Yamamoto, Ramaswamy, Lindsay, Lindsay, Feng, Lin, Zha, Patil, Shankar, Zhang, Zhang, Wang, Agarwal, Sajuyigbe, Chintala, Max, Chen, Kehoe, Satterfield, Govindaprasad, Gupta, Deng, Cho, Virk, Subramanian, Choudhury,
  Goldman, Remez, Glaser, Best, Koehler, Robinson, Li, Zhang, Matthews, Chou, Shaked, Vontimitta, Ajayi, Montanez, Mohan, Kumar, Mangla, Ionescu, Poenaru, Mihailescu, Ivanov, Li, Wang, Jiang, Bouaziz, Constable, Tang, Wu, Wang, Wu, Gao, Kleinman, Chen, Hu, Jia, Qi, Li, Zhang, Zhang, Adi, Nam, Yu, Wang, Zhao, Hao, Qian, Li, He, Rait, DeVito, Rosnbrick, Wen, Yang, Zhao, and Ma}]{grattafiori2024llama3herdmodels}
Aaron Grattafiori, Abhimanyu Dubey, Abhinav Jauhri, Abhinav Pandey, Abhishek Kadian, Ahmad Al-Dahle, Aiesha Letman, Akhil Mathur, Alan Schelten, Alex Vaughan, Amy Yang, Angela Fan, Anirudh Goyal, Anthony Hartshorn, Aobo Yang, Archi Mitra, Archie Sravankumar, Artem Korenev, Arthur Hinsvark, and 542 others. 2024.
\newblock \href {https://arxiv.org/abs/2407.21783} {The llama 3 herd of models}.
\newblock \emph{ArXiv preprint}, abs/2407.21783.

\bibitem[{Guo et~al.(2024{\natexlab{a}})Guo, Wang, Guo, Li, Song, Tan, Liu, Bian, and Yang}]{guo2024connecting}
Qingyan Guo, Rui Wang, Junliang Guo, Bei Li, Kaitao Song, Xu~Tan, Guoqing Liu, Jiang Bian, and Yujiu Yang. 2024{\natexlab{a}}.
\newblock \href {https://openreview.net/forum?id=ZG3RaNIsO8} {Connecting large language models with evolutionary algorithms yields powerful prompt optimizers}.
\newblock In \emph{The Twelfth International Conference on Learning Representations, {ICLR} 2024, Vienna, Austria, May 7-11, 2024}. OpenReview.net.

\bibitem[{Guo et~al.(2024{\natexlab{b}})Guo, Huang, Liu, Fan, V{\'e}lez, Wu, Wang, Griffiths, and Wang}]{guo2024embodied}
Xudong Guo, Kaixuan Huang, Jiale Liu, Wenhui Fan, Natalia V{\'e}lez, Qingyun Wu, Huazheng Wang, Thomas~L. Griffiths, and Mengdi Wang. 2024{\natexlab{b}}.
\newblock \href {https://openreview.net/forum?id=VKlrzygQlT} {Embodied {LLM} agents learn to cooperate in organized teams}.
\newblock In \emph{Language Gamification - NeurIPS 2024 Workshop}.

\bibitem[{Hao et~al.(2025)Hao, Hu, Qi, Wu, Zhang, and Nie}]{hao2023chatllmnetworkbrainsintelligence}
Rui Hao, Linmei Hu, Weijian Qi, Qingliu Wu, Yirui Zhang, and Liqiang Nie. 2025.
\newblock \href {https://doi.org/10.1016/j.aiopen.2025.01.001} {Chatllm network: More brains, more intelligence}.
\newblock \emph{AI Open}, 6:45--52.

\bibitem[{Hendrycks et~al.(2021{\natexlab{a}})Hendrycks, Burns, Basart, Zou, Mazeika, Song, and Steinhardt}]{hendrycks2021measuring}
Dan Hendrycks, Collin Burns, Steven Basart, Andy Zou, Mantas Mazeika, Dawn Song, and Jacob Steinhardt. 2021{\natexlab{a}}.
\newblock \href {https://openreview.net/forum?id=d7KBjmI3GmQ} {Measuring massive multitask language understanding}.
\newblock In \emph{9th International Conference on Learning Representations, {ICLR} 2021, Virtual Event, Austria, May 3-7, 2021}. OpenReview.net.

\bibitem[{Hendrycks et~al.(2021{\natexlab{b}})Hendrycks, Burns, Kadavath, Arora, Basart, Tang, Song, and Steinhardt}]{hendrycksmath2021}
Dan Hendrycks, Collin Burns, Saurav Kadavath, Akul Arora, Steven Basart, Eric Tang, Dawn Song, and Jacob Steinhardt. 2021{\natexlab{b}}.
\newblock Measuring mathematical problem solving with the math dataset.
\newblock \emph{NeurIPS}.

\bibitem[{Higgs et~al.(2005)Higgs, Plewnia, and Ploch}]{higgs2005influence}
Malcolm Higgs, Ulrich Plewnia, and Jorg Ploch. 2005.
\newblock \href {http://eprints.soton.ac.uk/id/eprint/51415} {Influence of team composition and task complexity on team performance}.
\newblock \emph{Team Performance Management: An International Journal}, 11(7/8):227--250.

\bibitem[{Hong et~al.(2024)Hong, Zhuge, Chen, Zheng, Cheng, Wang, Zhang, Wang, Yau, Lin, Zhou, Ran, Xiao, Wu, and Schmidhuber}]{hong2024metagpt}
Sirui Hong, Mingchen Zhuge, Jonathan Chen, Xiawu Zheng, Yuheng Cheng, Jinlin Wang, Ceyao Zhang, Zili Wang, Steven Ka~Shing Yau, Zijuan Lin, Liyang Zhou, Chenyu Ran, Lingfeng Xiao, Chenglin Wu, and J{\"{u}}rgen Schmidhuber. 2024.
\newblock \href {https://openreview.net/forum?id=VtmBAGCN7o} {Metagpt: Meta programming for {A} multi-agent collaborative framework}.
\newblock In \emph{The Twelfth International Conference on Learning Representations, {ICLR} 2024, Vienna, Austria, May 7-11, 2024}. OpenReview.net.

\bibitem[{Hu et~al.(2025)Hu, Lu, and Clune}]{hu2025automateddesignagenticsystems}
Shengran Hu, Cong Lu, and Jeff Clune. 2025.
\newblock \href {https://openreview.net/forum?id=t9U3LW7JVX} {Automated design of agentic systems}.
\newblock In \emph{The Thirteenth International Conference on Learning Representations}.

\bibitem[{Ishibashi and Nishimura(2024)}]{ishibashi2024selforganizedagentsllmmultiagent}
Yoichi Ishibashi and Yoshimasa Nishimura. 2024.
\newblock \href {https://arxiv.org/abs/2404.02183} {Self-organized agents: A llm multi-agent framework toward ultra large-scale code generation and optimization}.
\newblock \emph{ArXiv preprint}, abs/2404.02183.

\bibitem[{Khattab et~al.(2024)Khattab, Singhvi, Maheshwari, Zhang, Santhanam, Vardhamanan, Haq, Sharma, Joshi, Moazam, Miller, Zaharia, and Potts}]{khattab2024dspy}
Omar Khattab, Arnav Singhvi, Paridhi Maheshwari, Zhiyuan Zhang, Keshav Santhanam, Sri Vardhamanan, Saiful Haq, Ashutosh Sharma, Thomas~T. Joshi, Hanna Moazam, Heather Miller, Matei Zaharia, and Christopher Potts. 2024.
\newblock \href {https://openreview.net/forum?id=sY5N0zY5Od} {Dspy: Compiling declarative language model calls into state-of-the-art pipelines}.
\newblock In \emph{The Twelfth International Conference on Learning Representations, {ICLR} 2024, Vienna, Austria, May 7-11, 2024}. OpenReview.net.

\bibitem[{Kozlowski and Ilgen(2006)}]{kozlowski2006enhancing}
Steve~WJ Kozlowski and Daniel~R Ilgen. 2006.
\newblock \href {https://doi.org/10.1111/j.1529-1006.2006.00030.x} {Enhancing the effectiveness of work groups and teams}.
\newblock \emph{Psychological science in the public interest}, 7(3):77--124.

\bibitem[{Li et~al.(2023)Li, Hammoud, Itani, Khizbullin, and Ghanem}]{li2023camel}
Guohao Li, Hasan Hammoud, Hani Itani, Dmitrii Khizbullin, and Bernard Ghanem. 2023.
\newblock \href {http://papers.nips.cc/paper\_files/paper/2023/hash/a3621ee907def47c1b952ade25c67698-Abstract-Conference.html} {{CAMEL:} communicative agents for "mind" exploration of large language model society}.
\newblock In \emph{Advances in Neural Information Processing Systems 36: Annual Conference on Neural Information Processing Systems 2023, NeurIPS 2023, New Orleans, LA, USA, December 10 - 16, 2023}.

\bibitem[{Liang et~al.(2024)Liang, He, Jiao, Wang, Wang, Wang, Yang, Shi, and Tu}]{liang2024encouraging}
Tian Liang, Zhiwei He, Wenxiang Jiao, Xing Wang, Yan Wang, Rui Wang, Yujiu Yang, Shuming Shi, and Zhaopeng Tu. 2024.
\newblock \href {https://aclanthology.org/2024.emnlp-main.992} {Encouraging divergent thinking in large language models through multi-agent debate}.
\newblock In \emph{EMNLP}, pages 17889--17904.

\bibitem[{Lin et~al.(2023)Lin, Fu, Yang, Brahman, Huang, Bhagavatula, Ammanabrolu, Choi, and Ren}]{lin2023swiftsage}
Bill~Yuchen Lin, Yicheng Fu, Karina Yang, Faeze Brahman, Shiyu Huang, Chandra Bhagavatula, Prithviraj Ammanabrolu, Yejin Choi, and Xiang Ren. 2023.
\newblock \href {http://papers.nips.cc/paper\_files/paper/2023/hash/4b0eea69deea512c9e2c469187643dc2-Abstract-Conference.html} {Swiftsage: {A} generative agent with fast and slow thinking for complex interactive tasks}.
\newblock In \emph{Advances in Neural Information Processing Systems 36: Annual Conference on Neural Information Processing Systems 2023, NeurIPS 2023, New Orleans, LA, USA, December 10 - 16, 2023}.

\bibitem[{Ling et~al.(2017)Ling, Yogatama, Dyer, and Blunsom}]{ling2017program}
Wang Ling, Dani Yogatama, Chris Dyer, and Phil Blunsom. 2017.
\newblock \href {https://doi.org/10.18653/v1/P17-1015} {Program induction by rationale generation: Learning to solve and explain algebraic word problems}.
\newblock In \emph{Proceedings of the 55th Annual Meeting of the Association for Computational Linguistics (Volume 1: Long Papers)}, pages 158--167, Vancouver, Canada. Association for Computational Linguistics.

\bibitem[{Liu et~al.(2024{\natexlab{a}})Liu, Feng, Xue, Wang, Wu, Lu, Zhao, Deng, Zhang, Ruan et~al.}]{liu2024deepseek}
Aixin Liu, Bei Feng, Bing Xue, Bingxuan Wang, Bochao Wu, Chengda Lu, Chenggang Zhao, Chengqi Deng, Chenyu Zhang, Chong Ruan, and 1 others. 2024{\natexlab{a}}.
\newblock \href {https://arxiv.org/abs/2412.19437} {Deepseek-v3 technical report}.
\newblock \emph{ArXiv preprint}, abs/2412.19437.

\bibitem[{Liu et~al.(2024{\natexlab{b}})Liu, Zhang, Li, Liu, and Yang}]{liu2023dynamic}
Zijun Liu, Yanzhe Zhang, Peng Li, Yang Liu, and Diyi Yang. 2024{\natexlab{b}}.
\newblock \href {https://openreview.net/forum?id=XII0Wp1XA9} {A dynamic {LLM}-powered agent network for task-oriented agent collaboration}.
\newblock In \emph{First Conference on Language Modeling}.

\bibitem[{Nikoleizig et~al.(2019)Nikoleizig, Nestler, and Krause}]{nikoleizig2019prediction}
Lucie Nikoleizig, Steffen Nestler, and Sascha Krause. 2019.
\newblock \href {https://doi.org/10.1525/collabra.220} {Prediction of group performance: The interplay of individual performance, interpersonal attraction, and interpersonal behavior}.
\newblock \emph{Collabra: Psychology}, 5(1).

\bibitem[{OpenAI et~al.(2023)OpenAI, Achiam, Adler, Agarwal, Ahmad, Akkaya, Aleman, Almeida, Altenschmidt, Altman, Anadkat, Avila, Babuschkin, Balaji, Balcom, Baltescu, Bao, Bavarian, Belgum, Bello, Berdine, Bernadett-Shapiro, Berner, Bogdonoff, Boiko, Boyd, Brakman, Brockman, Brooks, Brundage, Button, Cai, Campbell, Cann, Carey, Carlson, Carmichael, Chan, Chang, Chantzis, Chen, Chen, Chen, Chen, Chen, Chess, Cho, Chu, Chung, Cummings, Currier, Dai, Decareaux, Degry, Deutsch, Deville, Dhar, Dohan, Dowling, Dunning, Ecoffet, Eleti, Eloundou, Farhi, Fedus, Felix, Fishman, Forte, Fulford, Gao, Georges, Gibson, Goel, Gogineni, Goh, Gontijo-Lopes, Gordon, Grafstein, Gray, Greene, Gross, Gu, Guo, Hallacy, Han, Harris, He, Heaton, Heidecke, Hesse, Hickey, Hickey, Hoeschele, Houghton, Hsu, Hu, Hu, Huizinga, Jain, Jain, Jang, Jiang, Jiang, Jin, Jin, Jomoto, Jonn, Jun, Kaftan, Łukasz Kaiser, Kamali, Kanitscheider, Keskar, Khan, Kilpatrick, Kim, Kim, Kim, Kirchner, Kiros, Knight, Kokotajlo, Łukasz Kondraciuk,
  Kondrich, Konstantinidis, Kosic, Krueger, Kuo, Lampe, Lan, Lee, Leike, Leung, Levy, Li, Lim, Lin, Lin, Litwin, Lopez, Lowe, Lue, Makanju, Malfacini, Manning, Markov, Markovski, Martin, Mayer, Mayne, McGrew, McKinney, McLeavey, McMillan, McNeil, Medina, Mehta, Menick, Metz, Mishchenko, Mishkin, Monaco, Morikawa, Mossing, Mu, Murati, Murk, Mély, Nair, Nakano, Nayak, Neelakantan, Ngo, Noh, Ouyang, O'Keefe, Pachocki, Paino, Palermo, Pantuliano, Parascandolo, Parish, Parparita, Passos, Pavlov, Peng, Perelman, de~Avila Belbute~Peres, Petrov, de~Oliveira~Pinto, Michael, Pokorny, Pokrass, Pong, Powell, Power, Power, Proehl, Puri, Radford, Rae, Ramesh, Raymond, Real, Rimbach, Ross, Rotsted, Roussez, Ryder, Saltarelli, Sanders, Santurkar, Sastry, Schmidt, Schnurr, Schulman, Selsam, Sheppard, Sherbakov, Shieh, Shoker, Shyam, Sidor, Sigler, Simens, Sitkin, Slama, Sohl, Sokolowsky, Song, Staudacher, Such, Summers, Sutskever, Tang, Tezak, Thompson, Tillet, Tootoonchian, Tseng, Tuggle, Turley, Tworek, Uribe, Vallone,
  Vijayvergiya, Voss, Wainwright, Wang, Wang, Wang, Ward, Wei, Weinmann, Welihinda, Welinder, Weng, Weng, Wiethoff, Willner, Winter, Wolrich, Wong, Workman, Wu, Wu, Wu, Xiao, Xu, Yoo, Yu, Yuan, Zaremba, Zellers, Zhang, Zhang, Zhao, Zheng, Zhuang, Zhuk, and Zoph}]{openai2024gpt4technicalreport}
OpenAI, Josh Achiam, Steven Adler, Sandhini Agarwal, Lama Ahmad, Ilge Akkaya, Florencia~Leoni Aleman, Diogo Almeida, Janko Altenschmidt, Sam Altman, Shyamal Anadkat, Red Avila, Igor Babuschkin, Suchir Balaji, Valerie Balcom, Paul Baltescu, Haiming Bao, Mohammad Bavarian, Jeff Belgum, and 262 others. 2023.
\newblock \href {https://arxiv.org/abs/2303.08774} {Gpt-4 technical report}.
\newblock \emph{ArXiv preprint}, abs/2303.08774.

\bibitem[{Pan et~al.(2025)Pan, Cemri, Agrawal, Yang, Chopra, Tiwari, Keutzer, Parameswaran, Ramchandran, Klein, Gonzalez, Zaharia, and Stoica}]{cemri2025multiagentllmsystemsfail}
Melissa~Z Pan, Mert Cemri, Lakshya~A Agrawal, Shuyi Yang, Bhavya Chopra, Rishabh Tiwari, Kurt Keutzer, Aditya Parameswaran, Kannan Ramchandran, Dan Klein, Joseph~E. Gonzalez, Matei Zaharia, and Ion Stoica. 2025.
\newblock \href {https://openreview.net/forum?id=wM521FqPvI} {Why do multiagent systems fail?}
\newblock In \emph{ICLR 2025 Workshop on Building Trust in Language Models and Applications}.

\bibitem[{Patel et~al.(2021)Patel, Bhattamishra, and Goyal}]{patel-etal-2021-nlp}
Arkil Patel, Satwik Bhattamishra, and Navin Goyal. 2021.
\newblock \href {https://doi.org/10.18653/v1/2021.naacl-main.168} {Are {NLP} models really able to solve simple math word problems?}
\newblock In \emph{Proceedings of the 2021 Conference of the North American Chapter of the Association for Computational Linguistics: Human Language Technologies}, pages 2080--2094, Online. Association for Computational Linguistics.

\bibitem[{Qian et~al.(2024)Qian, Liu, Liu, Chen, Dang, Li, Yang, Chen, Su, Cong, Xu, Li, Liu, and Sun}]{qian-etal-2024-chatdev}
Chen Qian, Wei Liu, Hongzhang Liu, Nuo Chen, Yufan Dang, Jiahao Li, Cheng Yang, Weize Chen, Yusheng Su, Xin Cong, Juyuan Xu, Dahai Li, Zhiyuan Liu, and Maosong Sun. 2024.
\newblock \href {https://doi.org/10.18653/v1/2024.acl-long.810} {{C}hat{D}ev: Communicative agents for software development}.
\newblock In \emph{Proceedings of the 62nd Annual Meeting of the Association for Computational Linguistics (Volume 1: Long Papers)}, pages 15174--15186, Bangkok, Thailand. Association for Computational Linguistics.

\bibitem[{Reimers and Gurevych(2019)}]{reimers2019sentence}
Nils Reimers and Iryna Gurevych. 2019.
\newblock \href {https://doi.org/10.18653/v1/D19-1410} {Sentence-{BERT}: Sentence embeddings using {S}iamese {BERT}-networks}.
\newblock In \emph{Proceedings of the 2019 Conference on Empirical Methods in Natural Language Processing and the 9th International Joint Conference on Natural Language Processing (EMNLP-IJCNLP)}, pages 3982--3992, Hong Kong, China. Association for Computational Linguistics.

\bibitem[{Roy and Roth(2015)}]{roy-roth-2015-solving}
Subhro Roy and Dan Roth. 2015.
\newblock \href {https://doi.org/10.18653/v1/D15-1202} {Solving general arithmetic word problems}.
\newblock In \emph{Proceedings of the 2015 Conference on Empirical Methods in Natural Language Processing}, pages 1743--1752, Lisbon, Portugal. Association for Computational Linguistics.

\bibitem[{Song et~al.(2025)Song, Liu, Zhang, Zhang, Luo, Wang, Wu, and Wang}]{song2025adaptiveinconversationteambuilding}
Linxin Song, Jiale Liu, Jieyu Zhang, Shaokun Zhang, Ao~Luo, Shijian Wang, Qingyun Wu, and Chi Wang. 2025.
\newblock \href {https://arxiv.org/abs/2405.19425} {Adaptive in-conversation team building for language model agents}.
\newblock \emph{Preprint}, arXiv:2405.19425.

\bibitem[{Suzgun and Kalai(2024)}]{suzgun2024metapromptingenhancinglanguagemodels}
Mirac Suzgun and Adam~Tauman Kalai. 2024.
\newblock \href {https://arxiv.org/abs/2401.12954} {Meta-prompting: Enhancing language models with task-agnostic scaffolding}.
\newblock \emph{Preprint}, arXiv:2401.12954.

\bibitem[{Tam et~al.(2024)Tam, Wu, Tsai, Lin, Lee, and Chen}]{tam2024letspeakfreelystudy}
Zhi~Rui Tam, Cheng-Kuang Wu, Yi-Lin Tsai, Chieh-Yen Lin, Hung-yi Lee, and Yun-Nung Chen. 2024.
\newblock \href {https://doi.org/10.18653/v1/2024.emnlp-industry.91} {Let me speak freely? a study on the impact of format restrictions on large language model performance.}
\newblock In \emph{Proceedings of the 2024 Conference on Empirical Methods in Natural Language Processing: Industry Track}, pages 1218--1236, Miami, Florida, US. Association for Computational Linguistics.

\bibitem[{Team et~al.(2024)Team, Georgiev, Lei, Burnell, Bai, Gulati, Tanzer, Vincent, Pan, Wang et~al.}]{team2024gemini}
Gemini Team, Petko Georgiev, Ving~Ian Lei, Ryan Burnell, Libin Bai, Anmol Gulati, Garrett Tanzer, Damien Vincent, Zhufeng Pan, Shibo Wang, and 1 others. 2024.
\newblock \href {https://arxiv.org/abs/2403.05530} {Gemini 1.5: Unlocking multimodal understanding across millions of tokens of context}.
\newblock \emph{ArXiv preprint}, abs/2403.05530.

\bibitem[{Wang et~al.(2025{\natexlab{a}})Wang, Wang, Athiwaratkun, Zhang, and Zou}]{wang2025mixtureofagents}
Junlin Wang, Jue Wang, Ben Athiwaratkun, Ce~Zhang, and James Zou. 2025{\natexlab{a}}.
\newblock \href {https://openreview.net/forum?id=h0ZfDIrj7T} {Mixture-of-agents enhances large language model capabilities}.
\newblock In \emph{The Thirteenth International Conference on Learning Representations}.

\bibitem[{Wang et~al.(2024{\natexlab{a}})Wang, Ma, Feng, Zhang, Yang, Zhang, Chen, Tang, Chen, Lin, Zhao, Wei, and Wen}]{Wang_2024}
Lei Wang, Chen Ma, Xueyang Feng, Zeyu Zhang, Hao Yang, Jingsen Zhang, Zhiyuan Chen, Jiakai Tang, Xu~Chen, Yankai Lin, Wayne~Xin Zhao, Zhewei Wei, and Jirong Wen. 2024{\natexlab{a}}.
\newblock \href {https://doi.org/10.1007/s11704-024-40231-1} {A survey on large language model based autonomous agents}.
\newblock \emph{Frontiers Comput. Sci.}, 18(6):186345.

\bibitem[{Wang et~al.(2023)Wang, Li, Chen, Cai, Zhu, Lin, Cao, Liu, Liu, and Sui}]{wang2023largelanguagemodelsfair}
Peiyi Wang, Lei Li, Liang Chen, Zefan Cai, Dawei Zhu, Binghuai Lin, Yunbo Cao, Qi~Liu, Tianyu Liu, and Zhifang Sui. 2023.
\newblock \href {https://arxiv.org/abs/2305.17926} {Large language models are not fair evaluators}.
\newblock \emph{ArXiv preprint}, abs/2305.17926.

\bibitem[{Wang et~al.(2022)Wang, Jansen, C{\^o}t{\'e}, and Ammanabrolu}]{wang2022scienceworld}
Ruoyao Wang, Peter Jansen, Marc-Alexandre C{\^o}t{\'e}, and Prithviraj Ammanabrolu. 2022.
\newblock \href {https://doi.org/10.18653/v1/2022.emnlp-main.775} {{S}cience{W}orld: Is your agent smarter than a 5th grader?}
\newblock In \emph{Proceedings of the 2022 Conference on Empirical Methods in Natural Language Processing}, pages 11279--11298, Abu Dhabi, United Arab Emirates. Association for Computational Linguistics.

\bibitem[{Wang et~al.(2020)Wang, Wei, Dong, Bao, Yang, and Zhou}]{wang2020minilm}
Wenhui Wang, Furu Wei, Li~Dong, Hangbo Bao, Nan Yang, and Ming Zhou. 2020.
\newblock \href {https://proceedings.neurips.cc/paper/2020/hash/3f5ee243547dee91fbd053c1c4a845aa-Abstract.html} {Minilm: Deep self-attention distillation for task-agnostic compression of pre-trained transformers}.
\newblock In \emph{Advances in Neural Information Processing Systems 33: Annual Conference on Neural Information Processing Systems 2020, NeurIPS 2020, December 6-12, 2020, virtual}.

\bibitem[{Wang et~al.(2025{\natexlab{b}})Wang, Zeng, Liu, Wong, Meng, Zhou, and Zhang}]{wang2025delta}
Yutong Wang, Jiali Zeng, Xuebo Liu, Derek~F. Wong, Fandong Meng, Jie Zhou, and Min Zhang. 2025{\natexlab{b}}.
\newblock \href {https://openreview.net/forum?id=hoYFLRNbhc} {Del{TA}: An online document-level translation agent based on multi-level memory}.
\newblock In \emph{The Thirteenth International Conference on Learning Representations}.

\bibitem[{Wang et~al.(2024{\natexlab{b}})Wang, Mao, Wu, Ge, Wei, and Ji}]{wang2024unleashingemergentcognitivesynergy}
Zhenhailong Wang, Shaoguang Mao, Wenshan Wu, Tao Ge, Furu Wei, and Heng Ji. 2024{\natexlab{b}}.
\newblock \href {https://aclanthology.org/2024.naacl-long.15} {Unleashing the emergent cognitive synergy in large language models: A task-solving agent through multi-persona self-collaboration}.
\newblock In \emph{Proceedings of the 2024 Conference of the North American Chapter of the Association for Computational Linguistics: Human Language Technologies (Volume 1: Long Papers)}, pages 257--279, Mexico City, Mexico. Association for Computational Linguistics.

\bibitem[{Wang et~al.(2025{\natexlab{c}})Wang, Wang, Liu, Ding, Zhang, Liu, and Zhang}]{wang2025agentdropoutdynamicagentelimination}
Zhexuan Wang, Yutong Wang, Xuebo Liu, Liang Ding, Miao Zhang, Jie Liu, and Min Zhang. 2025{\natexlab{c}}.
\newblock \href {https://arxiv.org/abs/2503.18891} {Agentdropout: Dynamic agent elimination for token-efficient and high-performance llm-based multi-agent collaboration}.
\newblock \emph{ArXiv preprint}, abs/2503.18891.

\bibitem[{Wataoka et~al.(2024)Wataoka, Takahashi, and Ri}]{wataoka2024selfpreferencebiasllmasajudge}
Koki Wataoka, Tsubasa Takahashi, and Ryokan Ri. 2024.
\newblock \href {https://arxiv.org/abs/2410.21819} {Self-preference bias in llm-as-a-judge}.
\newblock \emph{ArXiv preprint}, abs/2410.21819.

\bibitem[{Wei et~al.(2022)Wei, Wang, Schuurmans, Bosma, Ichter, Xia, Chi, Le, and Zhou}]{cot}
Jason Wei, Xuezhi Wang, Dale Schuurmans, Maarten Bosma, Brian Ichter, Fei Xia, Ed~H. Chi, Quoc~V. Le, and Denny Zhou. 2022.
\newblock \href {http://papers.nips.cc/paper\_files/paper/2022/hash/9d5609613524ecf4f15af0f7b31abca4-Abstract-Conference.html} {Chain-of-thought prompting elicits reasoning in large language models}.
\newblock In \emph{Advances in Neural Information Processing Systems 35: Annual Conference on Neural Information Processing Systems 2022, NeurIPS 2022, New Orleans, LA, USA, November 28 - December 9, 2022}.

\bibitem[{Wu et~al.(2024)Wu, Bansal, Zhang, Wu, Li, Zhu, Jiang, Zhang, Zhang, Liu, Awadallah, White, Burger, and Wang}]{wu2023autogenenablingnextgenllm}
Qingyun Wu, Gagan Bansal, Jieyu Zhang, Yiran Wu, Beibin Li, Erkang Zhu, Li~Jiang, Xiaoyun Zhang, Shaokun Zhang, Jiale Liu, Ahmed~Hassan Awadallah, Ryen~W White, Doug Burger, and Chi Wang. 2024.
\newblock \href {https://openreview.net/forum?id=BAakY1hNKS} {Autogen: Enabling next-gen {LLM} applications via multi-agent conversations}.
\newblock In \emph{First Conference on Language Modeling}.

\bibitem[{Xu et~al.(2023)Xu, Yang, Lin, Wang, Zhou, Zhang, and Mao}]{xu2025expertpromptinginstructinglargelanguage}
Benfeng Xu, An~Yang, Junyang Lin, Quan Wang, Chang Zhou, Yongdong Zhang, and Zhendong Mao. 2023.
\newblock \href {https://arxiv.org/abs/2305.14688} {Expertprompting: Instructing large language models to be distinguished experts}.
\newblock \emph{ArXiv preprint}, abs/2305.14688.

\bibitem[{Yang et~al.(2024)Yang, Yang, Zhang, Hui, Zheng, Yu, Li, Liu, Huang, Wei, Lin, Yang, Tu, Zhang, Yang, Yang, Zhou, Lin, Dang, Lu, Bao, Yang, Yu, Li, Xue, Zhang, Zhu, Men, Lin, Li, Xia, Ren, Ren, Fan, Su, Zhang, Wan, Liu, Cui, Zhang, and Qiu}]{qwen2.5}
An~Yang, Baosong Yang, Beichen Zhang, Binyuan Hui, Bo~Zheng, Bowen Yu, Chengyuan Li, Dayiheng Liu, Fei Huang, Haoran Wei, Huan Lin, Jian Yang, Jianhong Tu, Jianwei Zhang, Jianxin Yang, Jiaxi Yang, Jingren Zhou, Junyang Lin, Kai Dang, and 22 others. 2024.
\newblock \href {https://arxiv.org/abs/2412.15115} {Qwen2.5 technical report}.
\newblock \emph{ArXiv preprint}, abs/2412.15115.

\bibitem[{Yao et~al.(2023)Yao, Zhao, Yu, Du, Shafran, Narasimhan, and Cao}]{yao2023react}
Shunyu Yao, Jeffrey Zhao, Dian Yu, Nan Du, Izhak Shafran, Karthik~R. Narasimhan, and Yuan Cao. 2023.
\newblock \href {https://openreview.net/pdf?id=WE\_vluYUL-X} {React: Synergizing reasoning and acting in language models}.
\newblock In \emph{The Eleventh International Conference on Learning Representations, {ICLR} 2023, Kigali, Rwanda, May 1-5, 2023}. OpenReview.net.

\bibitem[{Yuan et~al.(2025)Yuan, Song, Chen, Tan, Li, and Yang}]{yuan2025evoagentautomaticmultiagentgeneration}
Siyu Yuan, Kaitao Song, Jiangjie Chen, Xu~Tan, Dongsheng Li, and Deqing Yang. 2025.
\newblock \href {https://aclanthology.org/2025.naacl-long.315/} {{E}vo{A}gent: Towards automatic multi-agent generation via evolutionary algorithms}.
\newblock In \emph{Proceedings of the 2025 Conference of the Nations of the Americas Chapter of the Association for Computational Linguistics: Human Language Technologies (Volume 1: Long Papers)}, pages 6192--6217, Albuquerque, New Mexico. Association for Computational Linguistics.

\bibitem[{Zhang et~al.(2025{\natexlab{a}})Zhang, Chen, Wan, Chang, Cheng, Wang, Hu, and Bai}]{zhang2025evoflowevolvingdiverseagentic}
Guibin Zhang, Kaijie Chen, Guancheng Wan, Heng Chang, Hong Cheng, Kun Wang, Shuyue Hu, and Lei Bai. 2025{\natexlab{a}}.
\newblock \href {https://arxiv.org/abs/2502.07373} {Evoflow: Evolving diverse agentic workflows on the fly}.
\newblock \emph{ArXiv preprint}, abs/2502.07373.

\bibitem[{Zhang et~al.(2025{\natexlab{b}})Zhang, Yue, Li, Yun, Wan, Wang, Cheng, Yu, and Chen}]{zhang2025cut}
Guibin Zhang, Yanwei Yue, Zhixun Li, Sukwon Yun, Guancheng Wan, Kun Wang, Dawei Cheng, Jeffrey~Xu Yu, and Tianlong Chen. 2025{\natexlab{b}}.
\newblock \href {https://openreview.net/forum?id=LkzuPorQ5L} {Cut the crap: An economical communication pipeline for {LLM}-based multi-agent systems}.
\newblock In \emph{The Thirteenth International Conference on Learning Representations}.

\bibitem[{Zhang et~al.(2025{\natexlab{c}})Zhang, Xiang, Yu, Teng, Chen, Chen, Zhuge, Cheng, Hong, Wang, Zheng, Liu, Luo, and Wu}]{zhang2025aflowautomatingagenticworkflow}
Jiayi Zhang, Jinyu Xiang, Zhaoyang Yu, Fengwei Teng, Xiong-Hui Chen, Jiaqi Chen, Mingchen Zhuge, Xin Cheng, Sirui Hong, Jinlin Wang, Bingnan Zheng, Bang Liu, Yuyu Luo, and Chenglin Wu. 2025{\natexlab{c}}.
\newblock \href {https://openreview.net/forum?id=z5uVAKwmjf} {{AF}low: Automating agentic workflow generation}.
\newblock In \emph{The Thirteenth International Conference on Learning Representations}.

\bibitem[{Zheng et~al.(2023)Zheng, Chiang, Sheng, Zhuang, Wu, Zhuang, Lin, Li, Li, Xing, Zhang, Gonzalez, and Stoica}]{zheng2023judgingllmasajudgemtbenchchatbot}
Lianmin Zheng, Wei{-}Lin Chiang, Ying Sheng, Siyuan Zhuang, Zhanghao Wu, Yonghao Zhuang, Zi~Lin, Zhuohan Li, Dacheng Li, Eric~P. Xing, Hao Zhang, Joseph~E. Gonzalez, and Ion Stoica. 2023.
\newblock \href {http://papers.nips.cc/paper\_files/paper/2023/hash/91f18a1287b398d378ef22505bf41832-Abstract-Datasets\_and\_Benchmarks.html} {Judging llm-as-a-judge with mt-bench and chatbot arena}.
\newblock In \emph{Advances in Neural Information Processing Systems 36: Annual Conference on Neural Information Processing Systems 2023, NeurIPS 2023, New Orleans, LA, USA, December 10 - 16, 2023}.

\bibitem[{Zhuge et~al.(2024)Zhuge, Wang, Kirsch, Faccio, Khizbullin, and Schmidhuber}]{zhuge2024gptswarm}
Mingchen Zhuge, Wenyi Wang, Louis Kirsch, Francesco Faccio, Dmitrii Khizbullin, and J{\"{u}}rgen Schmidhuber. 2024.
\newblock \href {https://openreview.net/forum?id=uTC9AFXIhg} {Gptswarm: Language agents as optimizable graphs}.
\newblock In \emph{Forty-first International Conference on Machine Learning, {ICML} 2024, Vienna, Austria, July 21-27, 2024}. OpenReview.net.

\end{thebibliography}

\appendix

\section{Appendix}
\subsection{Pareto Criterion Overview}
\label{sec:Pareto}

In problems with multiple conflicting objectives, improving one objective often requires sacrificing performance on another. Therefore, a single optimal solution balancing all objectives usually does not exist. The Pareto criterion provides a principled approach to handle such trade-offs.

\paragraph{Dominance}  
Assuming both objectives (e.g., task relevance \( f_1 \) and diversity \( f_2 \)) are to be maximized, given two candidate solutions (teams) \( A_1 \) and \( A_2 \), we say \( A_1 \) \textit{dominates} \( A_2 \) (denoted \( A_1 \succ A_2 \)) if  
\[
f_1(A_1) \geq f_1(A_2), \quad f_2(A_1) \geq f_2(A_2),
\]
with at least one strict inequality.

\paragraph{Pareto Optimal Set}  
The set of all solutions not dominated by any other in the candidate set \( \mathcal{T} \) is called the \textit{Pareto optimal set}:  
\[
\mathcal{T}^* = \{ A \in \mathcal{T} \mid \nexists A' \in \mathcal{T}, \; A' \succ A \}.
\]

\paragraph{Pareto Front}  
Mapping solutions in the Pareto optimal set to the objective space forms the \textit{Pareto front}, which visually represents the best possible trade-offs between the objectives.

\paragraph{Application in AgentInit}  
We apply the Pareto criterion to select teams that achieve a reasonable trade-off between task relevance and diversity. By choosing from the Pareto optimal set, we ensure that the selected teams balance relevance and diversity, avoid arbitrary weighting of objectives, and support flexible selection strategies.

This approach effectively balances competing goals, resulting in diverse and high-performing agent teams for collaborative inference.

\subsection{Algorithm of AgentInit}
\label{sec:appendix}
We present the pseudocode of AgentInit in Algorithm~\ref{algo:full_process}. 

\subsection{Prompt Repository}

\begin{figure*}[h]
    \centering
    \begin{tcolorbox}[notitle, sharp corners, breakable, colframe=softblue, colback=white, 
       boxrule=3pt, boxsep=0.5pt, enhanced, 
       shadow={3pt}{-3pt}{0pt}{opacity=1,mygrey},
       title={\includegraphics[height=25pt]{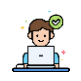} \large Selector}]
{\fontfamily{pcr}\selectfont\small
\begin{lstlisting}[breaklines=true]
'''
You are tasked with selecting the best group of experts to help answer a given question. Consider the relevance and effectiveness of each group's composition. 

# Question or Task
{context}

# Groups
{groups}

Follow these steps:
1. Analyze the Question: Identify its scope and complexity.
2. Evaluate Each Group: Assess the relevance of the roles in each group to the question. Compare the strengths and weaknesses of each group.
3. Make a Justified Choice: Select the most suitable group based on the above analysis.

# Attention 
1. A larger group is not necessarily better-some roles may be redundant or even negatively impact the results if they are not relevant to the question. Conversely, if the question requires a broad knowledge base, a more diverse group may be advantageous.
2. The last line of your response MUST be 'Choice: Group X' where X is the number of the selected group. For example, 'Choice: Group 1'.

'''
\end{lstlisting}
}
\end{tcolorbox}
    \caption{Prompt of pre-difined agents.}
    \label{fig:prompt}
\end{figure*}

We list the prompts used for different agents in the Standardized Agent Generation in Figure~\ref{fig:prompt}. Specifically, prompts for the \textit{Planner}, \textit{Observer}, \textit{Formatter}, and \textit{Selector} are provided.

\label{sec:prompt}
\subsection{Experiments on More Challenging Tasks}
\label{sec:science}
\subsubsection{Experimental Setup}
We conduct experiments on advanced mathematics problems using MATH~\citep{hendrycksmath2021} and AIME2025, as well as on Trivia Creative Writing~\citep{wang2024unleashingemergentcognitivesynergy} and ScienceWorld~\citep{wang2022scienceworld}, largely following the experimental setup and inference framework established by EvoAgent. The backbone model is replaced with Deepseek V3, and AutoAgents are incorporated into the baseline methods for comprehensive comparison.

\begin{table}[t]
\centering
\scalebox{0.9}{
\begin{tabular}{lrr}
\toprule
\textbf{Method} & \textbf{MATH} & \textbf{AIME2025} \\
\midrule
Vanilla      & 82.6 & 33.3 \\
CoT          & 80.2 & 36.7 \\
AgentPrune   & 84.0 & 36.7 \\
MASnone      & 79.6 & 31.5 \\
Pre-defined  & 83.2 & 23.3 \\
EvoAgent     & 83.8 & 40.0 \\
AutoAgents   & 83.3 & 42.2 \\
AgentInit    & \textbf{84.8} & \textbf{45.6} \\
\bottomrule
\end{tabular}
}
\caption{Performance comparison on advanced mathematics benchmarks (MATH and AIME2025) using Deepseek.}
\label{tab:appendix_math}
\end{table}

\begin{table}[t]
\centering
  \scalebox{0.75}{
\begin{tabular}{l|cc|cc}
\hline
\multirow{2}{*}{\textbf{Methods}} & \multicolumn{2}{c|}{\textbf{N = 5}} & \multicolumn{2}{c}{\textbf{N = 10}} \\
\cline{2-5}
 & \textbf{Score (\%)} & $\Delta$ & \textbf{Score (\%)} & $\Delta$ \\
\hline
Vanilla   & 74.2 & 0.0\% & 79.4 & 0.0\% \\
CoT       & 75.4 & \textcolor{teal!70!black}{+1.2\%} & 79.5 & \textcolor{teal!70!black}{+0.1\%} \\
Refine    & 78.8 & \textcolor{teal!70!black}{+4.6\%} & 81.5 & \textcolor{teal!70!black}{+2.1\%} \\
SPP       & 69.0 & \textcolor{red!80!black}{-5.2\%} & 65.2 & \textcolor{red!80!black}{-14.2\%} \\
EvoAgent$_{(1,3)}$  & 76.4 & \textcolor{teal!70!black}{+2.2\%} & 78.3 & \textcolor{red!80!black}{-1.1\%} \\
AutoAgents  & 78.6 & \textcolor{teal!70!black}{+4.2\%} & 80.0 &  \textcolor{teal!70!black}{+0.6\%}\\
AgentInit      & \textbf{79.2} & \textcolor{teal!70!black}{\textbf{+5.0\%}} & \textbf{81.8} & \textcolor{teal!70!black}{\textbf{+2.4\%}} \\
\hline
\end{tabular}
}
\caption{Trivia Creative Writing results on DeepSeek model. $\Delta$ indicates the score difference compared to the Vanilla baseline, and \textbf{N} denotes the number of questions associated with each story.}
\label{tab:writing-results}
\end{table} 
\subsubsection{Main Result}
\paragraph{Advanced Mathematics Tasks}
As shown in Table~\ref{tab:appendix_math}, AgentInit achieves the best performance on both datasets. On MATH, it surpasses the baselines, demonstrating improved reasoning on complex mathematics problems. On AIME2025, AgentInit outperforms the vanilla baseline by 12.3 points and the previous SOTA by 3.4 points, highlighting its strong effectiveness on more difficult, competition-level math tasks. These results supplement our evaluation on standard benchmarks and further underscore the advantage of AgentInit in challenging mathematical reasoning scenarios.

\paragraph{Trivia Creative Writing}
The results on the Trivia Creative Writing task, presented in Table~\ref{tab:writing-results}, show that AgentInit outperforms existing methods under both \(N=5\) and \(N=10\) settings, highlighting its strong effectiveness on more open-ended datasets.
Additionally, some existing initialization methods perform poorly in this setup, possibly due to limited adaptation to open-source models, while AgentInit demonstrates better adaptability.
\begin{table}[t]
\centering
\scalebox{0.7}{
\begin{tabular}{lrrrr}
\toprule
\textbf{Task Type} & \textbf{Vanilla}& \textbf{AutoAgents} & \textbf{EvoAgent$_{(1,1)}$} & \textbf{AgentInit} \\
\midrule
1-1 (L)  &  4.44 &5.11&9.22 &9.67  \\
1-2 (L)  &  8.11 &8.67&  20.67&  8.11\\
1-3 (L)  &  7.00 &3.33&  5.89& 8.11  \\
1-4 (L)  &  0.33 &8.33& 8.33 &  7.67\\
2-1 (M)  &  5.40 &14.90&  8.20&  13.90\\
2-2 (M)  &  11.30&15.80&  5.30&  12.50\\
2-3 (L)  &  18.70&13.00&  20.20&  14.70\\
3-1 (S)  &  44.20&53.40&  55.40&  54.80\\
3-2 (M)  &  22.20&24.20&  32.80&  23.20\\
3-3 (M)  &  47.50&47.00&  56.50&  61.00\\
3-4 (M)  & 74.80 &80.00& 76.50& 75.10\\
4-1 (S)  &  22.60&25.90&  17.60&  19.30\\
4-2 (S)  &  72.60&65.00& 64.20&  79.90\\
4-3 (S)  &  47.50&44.20& 40.10 & 41.00 \\
4-4 (S)  &  24.30&24.20&  18.40& 23.50\\
5-1 (L)  &  8.10 &8.40&  7.50&  8.90\\
5-2 (L)  &  24.80&33.20&  35.00&  25.70\\
6-1 (M)  &  21.25&23.00& 24.00  &24.12  \\
6-2 (S)  &  28.89&27.78&  25.56&  28.89\\
6-3 (M)  &  11.78&15.78&  11.00&  9.11\\
7-1 (S)  &  57.50&50.00&  52.50&  45.00\\
7-2 (S)  &  55.00&50.00& 45.00&  47.50\\
7-3 (S)  &  39.90&39.90&  43.10&  39.90\\
8-1 (M)  &  21.20&21.20&  21.20&  21.20\\
8-2 (S)  &  7.00 &6.00&  7.00&  7.00\\
9-1 (L)  &  22.50&31.00&  29.00&  30.00\\
9-2 (L)  &  8.89 &10.00&  11.67&  10.56\\
9-3 (L)  &  12.50&2.50&  2.00&  2.00\\
10-1 (L) & 3.30  &1.60&  3.40&  11.70\\
10-2 (L) &  25.30&18.60&  14.50&  33.70\\
\midrule
\textbf{Short} & \textbf{39.95} &38.64& 36.89 & \underline{38.68} \\
\textbf{Medium} & 26.93 &\textbf{30.23}& 29.43 & \underline{30.02} \\
\textbf{Long}  & 12.00 &11.98& \underline{13.95} & \textbf{14.23} \\
\midrule
\textbf{Overall}  & 25.29 &\underline{25.73}& 25.72 & \textbf{26.59} \\
\bottomrule
\end{tabular}
}
\caption{Detailed performance comparison across different task types based on Oracle Trajectory Length \citep{lin2023swiftsage} on the ScienceWorld.}
\label{tab:scienceworld}
\end{table}

\paragraph{ScienceWorld}
The results in Table~\ref{tab:scienceworld} demonstrate that AgentInit can be effectively used to initialize the multi-agent system (MAS) for solving scientific problems in dynamic, open-world environments. Compared to previous methods, AgentInit achieves significant improvements in overall performance as well as in long-horizon tasks. These results indicate that our method enhances the capability of the system to tackle complex tasks with extended trajectories.

\subsection{Efficiency under Different $N_{max}$ Settings}
\label{sec:efficiency}
As shown in Table~\ref{tab:main_tokens}, the input/output token usage of AgentInit is comparable to that of AutoAgents in some cases. One reason for this similarity is that the maximum number of agents $N_{max}$ is set to a relatively small value (5) in the main experiments, leading to team sizes similar to AutoAgents and thus comparable computational and token costs. To further examine scalability, we conducted additional experiments on the MMLU dataset with the Qwen2.5-72B-Instruct model when $N_{max}=10$, which exceeds the number of generated agents. As shown in Table~\ref{tab:appendix_efficiency}, AgentInit improves efficiency by approximately 10\% compared to AutoAgents under this setting. These results confirm that the efficiency advantage of AgentInit becomes more pronounced as the maximum team size increases.

\begin{table}[t]
\centering
\scalebox{0.9}{
\begin{tabular}{llrr}
\toprule
\textbf{\shortstack{Team Size}} & \textbf{Method} & \textbf{Ptok.} & \textbf{Ctok.} \\
\midrule
\multirow{2}{*}{5}  & AutoAgents & 704K  & 261K \\
                    & AgentInit  & 706K  & 267K \\
\midrule
\multirow{2}{*}{10} & AutoAgents & 1089K & 367K \\
                    & AgentInit  & 959K  & 339K \\
\bottomrule
\end{tabular}
}
\caption{Token usage comparison between AutoAgents and AgentInit under different maximum team sizes ($N_{max}$) on the MMLU dataset with Qwen2.5-72B-Instruct.}
\label{tab:appendix_efficiency}
\end{table}

\subsection{Case Study}
\label{sec:case}
We select two representative multiple-choice questions from the MMLU dataset to illustrate the reasoning process facilitated by the MAS initialized by AgentInit. For each case, we present the generated agents along with their individual responses, as well as highlight the agents that were dynamically dropped out during the interaction. These examples demonstrate how AgentInit optimizes team composition. The results are shown in Figure~\ref{fig:case}.

\subsection{Performance Analysis of Qwen in Various Frameworks}
\label{sec:qwen}
As shown in Table~\ref{tab:qwen_main_differentF}, AgentInit consistently achieves the best overall performance across various frameworks on the Qwen2.5-72B model. In graph structures such as Chain, Star, and Layered, as well as the AutoGen framework, AgentInit outperforms state-of-the-art methods by up to 1.1 points and predefined methods by up to 1.2 points.

\onecolumn

\begin{tcolorbox}[
notitle, sharp corners, breakable, colframe=softgreen, colback=white, 
       boxrule=3pt, boxsep=0.5pt, enhanced, 
       shadow={3pt}{-3pt}{0pt}{opacity=1,mygrey},
  title={\includegraphics[height=22pt]{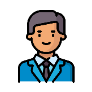} \large Planner}
]{\fontfamily{pcr}\selectfont\small
\begin{lstlisting}[breaklines=true]
'''
-----
You are a manager and an expert-level ChatGPT prompt engineer with expertise in multiple fields. Your goal is to break down tasks by creating exactly multiple LLM agents, assign them roles, analyze their dependencies, and provide a detailed execution plan. You should continuously improve the role list and plan based on the suggestions in the History section.

# Question or Task
{context}

# Existing Expert Roles
{existing_roles}

# History
{history}

# Steps
You will come up with solutions for any task or problem by following these steps:
1. You should first understand, analyze, and break down the human's problem/task.
2. According to the problem and existing expert roles, you will select the existing expert roles that are needed to solve the problem. You should act as an expert-level ChatGPT prompt engineer and planner with expertise in multiple fields, so that you can better develop a problem-solving plan and provide the best answer. You should follow these principles when selecting existing expert roles: 
2.1. Make full use of the existing expert roles to solve the problem. 
2.2. Follow the requirements of the existing expert roles. Make sure to select the existing expert roles that have cooperative or dependent relationships. 
3. According to the problem and existing expert roles, you will create additional expert roles that are needed to solve the problem. You should act as an expert-level ChatGPT prompt engineer and planner with expertise in multiple fields, so that you can better develop a problem-solving plan and provide the best answer. You should follow these principles when creating additional expert roles:
3.1. The newly created expert role should not have duplicate functions with any existing expert role. If there are duplicates, you do not need to create this role.
3.2. Each new expert role should include a name, a detailed description of their area of expertise, execution suggestions, and prompt templates.
3.3. Determine the number and domains of expertise of each new expert role based on the content of the problem. Please make sure each expert has a clear responsibility and do not let one expert do too many tasks. The description of their area of expertise should be detailed so that the role understands what they are capable of doing. 
3.4. Determine the names of each new expert role based on their domains of expertise. The name should express the characteristics of expert roles. 
3.5. Determine the goals of each new expert role based on their domains of expertise. The goal MUST indicate the primary responsibility or objective that the role aims to achieve. 
3.6. Determine the constraints of each new expert role based on their domains of expertise. The constraints MUST specify limitations or principles that the role must adhere to when performing actions. 
3.7. Provide some suggestions for each agent to execute the task, including but not limited to a clear output, extraction of historical information, and suggestions for execution steps. 
3.8. Generate the prompt template required for calling each new expert role according to its name, description, goal, constraints and suggestions.  A good prompt template should first explain the role it needs to play (name), its area of expertise (description), the primary responsibility or objective that the role aims to achieve (goal), limitations or principles that the role must adhere to when performing actions (constraints), and suggestions for agent to execute the task (suggestions). The prompt MUST follow the following format "You are [description], named [name]. Your goal is [goal], and your constraints are [constraints]. You could follow these execution suggestions: [suggestions].".
3.9. You MUST output the details of created new expert roles. Specifically, The new expert roles should have a `name` key (the expert role name), a `description` key (the description of the expert role's expertise domain), a `suggestions` key (some suggestions for each agent to execute the task), and a `prompt` key (the prompt template required to call the expert role).
4. Finally, based on the content of the problem/task and the expert roles, provide a detailed execution plan with the required steps to solve the problem.
4.1. The execution plan should consist of multiple steps that solve the problem progressively. Make the plan as detailed as possible to ensure the accuracy and completeness of the task. You need to make sure that the summary of all the steps can answer the question or complete the task.
4.2. Each step should assign one expert role to carry it out.
4.3. The description of each step should provide sufficient details and explain how the steps are connected to each other.
4.4. The description of each step must also include the expected output of that step and indicate what inputs are needed for the next step. The expected output of the current step and the required input for the next step must be consistent with each other. Sometimes, you may need to extract information or values before using them. Otherwise, the next step will lack the necessary input.
4.5. Output the execution plan as a numbered list of steps. For each step, please begin with a list of the expert roles that are involved in performing it.

# Suggestions
{suggestions}

# Attention
1. Please adhere to the requirements of the existing expert roles.
2. DO NOT answer the answer directly. You should focus on generating high-performance roles and a detailed plan to effectively address it.
3. If you do not receive any suggestions, you should always consider what kinds of expert roles are required and what are the essential steps to complete the tasks. If you do receive some suggestions, you should always evaluate how to enhance the previous role list and the execution plan according to these suggestions and what feedback you can give to the suggesters.
-----
'''

\end{lstlisting}
}
\end{tcolorbox}

\begin{tcolorbox}[notitle, sharp corners, breakable, colframe=softpink, colback=white, 
       boxrule=3pt, boxsep=0.5pt, enhanced, 
       shadow={3pt}{-3pt}{0pt}{opacity=1,mygrey},
       title={\includegraphics[height=25pt]{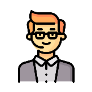}  \large Formatter}]
       {\fontfamily{pcr}\selectfont\small
       \begin{lstlisting}[breaklines=true]
'''
-----
You are a formatting expert. I will provide you with an agent planner's task execution plan, and you must strictly follow the requirements below. Extract the corresponding information and present it exactly in the specified format.
# Content to Format:
{raw_content}


# Format Requirements
1. Organize content into these sections:
   - Selected Roles List (JSON blobs)
   - Created Roles List (JSON blobs) 
   - Execution Plan (numbered list) For each step, begin with a list of the expert roles involved in performing it.
   - RoleFeedback (feedback on the historical Role suggestions)
   - PlanFeedback (feedback on the historical Plan suggestions)

2. Your final output should ALWAYS in the following format:
{format_example}

3. Use '##' for section headers
4. Ensure all expert roles are properly formatted. Each JSON blob should only contain one expert role, and do NOT return a list of multiple expert roles. Here is an example of a valid JSON blob:
{{{{
    "name": "ROLE NAME",
    "description": "ROLE DESCRIPTONS",
    "suggestions": "EXECUTION SUGGESTIONS",
    "prompt": "ROLE PROMPT",
}}}}
5. The prompt field should start with "You are xxx"
-----
'''
'''
---
## Question or Task:
the input question you must answer / the input task you must finish

## Selected Roles List:
```
JSON BLOB 1,
JSON BLOB 2,
JSON BLOB 3
```

## Created Roles List:
```
JSON BLOB 1,
JSON BLOB 2,
JSON BLOB 3
```

## Execution Plan:
1. [ROLE 1, ROLE2, ...]: STEP 1
2. [ROLE 1, ROLE2, ...]: STEP 2
2. [ROLE 1, ROLE2, ...]: STEP 3

## RoleFeedback
feedback on the historical Role suggestions

## PlanFeedback
feedback on the historical Plan suggestions
---
'''
\end{lstlisting}
}

\end{tcolorbox}

\begin{tcolorbox}[notitle, sharp corners, breakable, colframe=softpurple, colback=white, 
       boxrule=3pt, boxsep=0.5pt, enhanced, 
       shadow={3pt}{-3pt}{0pt}{opacity=1,mygrey},
       title={\includegraphics[height=25pt]{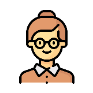} \large Observer}]
       {\fontfamily{pcr}\selectfont\small
\begin{lstlisting}[breaklines=true]
'''
-----
You are a ChatGPT executive observer expert skilled in identifying problem-solving plans and errors in the execution process. Your goal is to check if the created Expert Roles following the requirements and give your improvement suggestions. You can refer to historical suggestions in the History section, but try not to repeat them.

# Question or Task
{question}

# Existing Expert Roles
{existing_roles}

# Selected Roles List
{selected_roles}

# Created Roles List
{created_roles}

# History
{history}

# Steps
You will check the selected roles list and created roles list by following these steps:
1. You should first understand, analyze, and break down the human's problem/task.
2. According to the problem and existing expert roles, you should check the selected expert roles.
2.1. You should make sure that the selected expert roles can help you solve the problem effectively and efficiently.
2.2. You should make sure that the selected expert roles meet the requirements of the problem and have cooperative or dependent relationships with each other. 
2.3. You should make sure that the JSON blob of each selected expert role contains its original information, such as name, description, and requirements.
3. According to the problem and existing expert roles, you should check the new expert roles that you have created.
3.1. You should avoid creating any new expert role that has duplicate functions with any existing expert role. If there are duplicates, you should use the existing expert role instead.
3.2. You should include the following information for each new expert role: a name, a detailed description of their area of expertise, some suggestions for executing the task, and a prompt template for calling them.
3.3. You should assign a clear and specific domain of expertise to each new expert role based on the content of the problem. You should not let one expert role do too many tasks or have vague responsibilities. The description of their area of expertise should be detailed enough to let them know what they are capable of doing. 
3.4. You should give a meaningful and expressive name to each new expert role based on their domain of expertise. The name should reflect the characteristics and functions of the expert role. 
3.5. You should state a clear and concise goal for each new expert role based on their domain of expertise. The goal must indicate the primary responsibility or objective that the expert role aims to achieve. 
3.6. You should specify any limitations or principles that each new expert role must adhere to when performing actions. These are called constraints and they must be consistent with the problem requirements and the domain of expertise. 
3.7. You should provide some helpful suggestions for each new expert role to execute the task effectively and efficiently. The suggestions should include but not limited to a clear output format, extraction of relevant information from previous steps, and guidance for execution steps.
3.8. You should create a prompt template for calling each new expert role according to its name, description, goal, constraints and suggestions. A good prompt template should first explain the role it needs to play (name), its area of expertise (description), the primary responsibility or objective that it aims to achieve (goal), any limitations or principles that it must adhere to when performing actions (constraints), and some helpful suggestions for executing the task (suggestions). The prompt must follow this format: "You are [description], named [name]. Your goal is [goal, and your constraints are [constraints]. You could follow these execution suggestions: [suggestions].".
3.9. You should follow the JSON blob format for creating new expert roles. Specifically, The JSON of new expert roles should have a `name` key (the expert role name), a `description` key (the description of the expert role's expertise domain), a `suggestions` key (some suggestions for each agent to execute the task), and a `prompt` key (the prompt template required to call the expert role). Each JSON blob should only contain one expert role, and do NOT return a list of multiple expert roles. Here is an example of a valid JSON blob:
{{{{
    "name": "ROLE NAME",
    "description": "ROLE DESCRIPTONS",
    "suggestions": "EXECUTION SUGGESTIONS",
    "prompt": "ROLE PROMPT",
}}}}
4. Output a summary of the inspection results above. If you find any errors or have any suggestions, please state them clearly in the Suggestions section. If there are no errors or suggestions, you MUST write 'No Suggestions' in the Suggestions section.

# Format example
Your final output should ALWAYS in the following format:
{format_example}

# Attention
1. Please adhere to the requirements of the existing expert roles.
2. You can refer to historical suggestions and feedback in the History section but DO NOT repeat historical suggestions.
3. DO NOT ask any questions to the user or human. 
-----
'''

FORMAT_EXAMPLE = '''
---
## Thought
you should always think about if there are any errors or suggestions for selected and created expert roles.

## Suggestions
1. ERROR1/SUGGESTION1
2. ERROR2/SUGGESTION2
2. ERROR3/SUGGESTION3
---

'''

'''
-----
You are a ChatGPT executive observer expert skilled in identifying problem-solving plans and errors in the execution process. Your goal is to check if the Execution Plan following the requirements and give your improvement suggestions. You can refer to historical suggestions in the History section, but try not to repeat them.

# Question or Task
{context}

# Role List
{roles}

# Execution Plan
{plan}

# History
{history}

# Steps
You will check the Execution Plan by following these steps:
1. You should first understand, analyze, and disassemble the human's problem.
2. You should check if the execution plan meets the following requirements:
2.1. The execution plan should consist of multiple steps that solve the problem progressively. Make the plan as detailed as possible to ensure the accuracy and completeness of the task. You need to make sure that the summary of all the steps can answer the question or complete the task.
2.2. Each step should assign at least one expert role to carry it out. If a step involves multiple expert roles, you need to specify the contributions of each expert role and how they collaborate to produce integrated results. 
2.3. The description of each step should provide sufficient details and explain how the steps are connected to each other.
2.4. The description of each step must also include the expected output of that step and indicate what inputs are needed for the next step. The expected output of the current step and the required input for the next step must be consistent with each other. Sometimes, you may need to extract information or values before using them. Otherwise, the next step will lack the necessary input.
3. Output a summary of the inspection results above. If you find any errors or have any suggestions, please state them clearly in the Suggestions section. If there are no errors or suggestions, you MUST write 'No Suggestions' in the Suggestions section.

# Format example
Your final output should ALWAYS in the following format:
{format_example}

# Attention
1. You can refer to historical suggestions and feedback in the History section but DO NOT repeat historical suggestions.
2. DO NOT ask any questions to the user or human.
-----
'''

FORMAT_EXAMPLE = '''
---
## Thought
you should always think about if there are any errors or suggestions for the Execution Plan.

## Suggestions
1. ERROR1/SUGGESTION1
2. ERROR2/SUGGESTION2
2. ERROR3/SUGGESTION3
---
'''

\end{lstlisting}
}
\end{tcolorbox}

\onecolumn

\begin{tcolorbox}[notitle, colback=red!5,colframe=softorange,
       boxrule=3pt, boxsep=0.5pt, enhanced, 
       shadow={3pt}{-3pt}{0pt}{opacity=1,mygrey},
       title={\includegraphics[height=22pt]{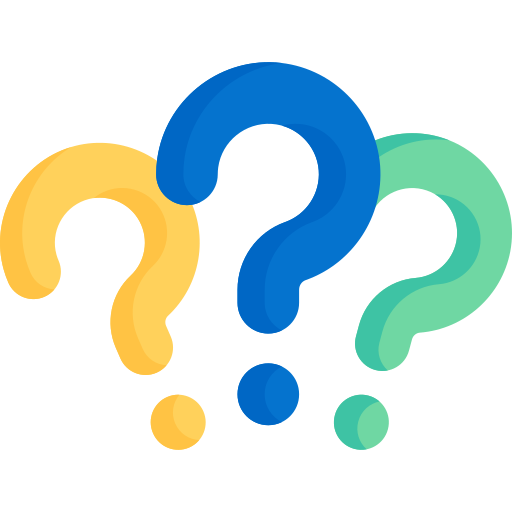} \LARGE Query}]
\texttt{
Fine argues that states are not clubs because\\
Option A: it is not possible today for would-be immigrants to get together to set up a state of their own.\\
Option B: exclusion from a golf club is unlikely to have a devastating impact on the life of a would-be member.\\
Option C: both A and B\\
Option D: neither A nor B
}
\end{tcolorbox}

\vspace{50pt}
\begin{tcolorbox}[notitle,breakable,colback=white,colframe=softpink,
       boxrule=3pt, boxsep=0.5pt, enhanced, 
       shadow={3pt}{-3pt}{0pt}{opacity=1,mygrey},
       title={\large (1/4)  \includegraphics[height=22pt, trim=0 100 0 0, clip]{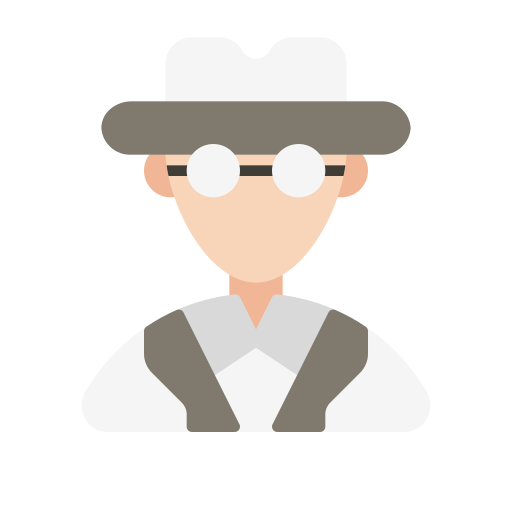}\large Comparative Sociologist}]
\texttt{
You are a Comparative Sociologist.\\ Your goal is to evaluate the social implications of being excluded from a state versus a club, and your constraints are to consider both historical and contemporary contexts.\\ You could follow these execution suggestions: Review the social and economic impacts of exclusion from both states and clubs, and provide a comparative analysis.\\
\rule[0mm]{\linewidth}{0.5pt}
To address this question, let's break down the arguments posed by Fine and evaluate the social and economic implications of exclusion from both states and clubs, drawing from both historical and contemporary contexts.\\ Option A: It is not possible today for would-be immigrants to get together to set up a state of their own.\\This statement highlights a fundamental difference between states and clubs.\\ States are sovereign entities that control a defined territory and have a monopoly on the legitimate use of force within that territory.\\ The process of forming a new state is complex and often requires international recognition, which is not easily achievable.\\ In contrast, clubs are voluntary associations that individuals can form or join relatively easily.\\ For example, a group of people can easily create a new golf club or social club, but they cannot create a new state without significant political and military resources.\\ Option B: Exclusion from a golf club is unlikely to have a devastating impact on the life of a would-be member.\\This statement underscores the differing levels of impact that exclusion from a state versus a club can have on an individual's life.\\ Exclusion from a state can lead to statelessness, which has severe social and economic consequences.\\ Stateless individuals often lack basic rights such as access to education, healthcare, and employment, and they may face significant legal and social barriers.\\ In contrast, exclusion from a club, such as a golf club, may be disappointing but is generally not life-altering.\\ While it can affect social status and networking opportunities, it does not fundamentally undermine one's ability to lead a normal life.\\ Comparative Analysis\\1.\\ Social and Economic Impacts of Exclusion from States:   - Statelessness: Being stateless can lead to a lack of legal protection, making individuals vulnerable to exploitation and abuse.\\   - Limited Access to Services: Stateless individuals often cannot access public services such as healthcare, education, and social welfare.\\   - Employment Barriers: Without legal status, finding employment can be extremely difficult, leading to unemployment or informal, precarious work.\\   - Social Isolation: Stateless individuals may face social isolation and exclusion from the broader community.\\2.\\ Social and Economic Impacts of Exclusion from Clubs:   - Social Status: Exclusion from a club can affect an individual's social status and standing within certain social circles.\\   - Networking: Clubs often provide networking opportunities that can be beneficial for career advancement and personal relationships.\\   - Recreational Activities: Exclusion from recreational clubs can limit access to enjoyable activities and social interactions.\\ Historical and Contemporary Contexts- Historical Context:  - Statelessness: Historically, statelessness has been a significant issue, particularly after events such as decolonization and the redrawing of national boundaries.\\ For example, the creation of new states in the Middle East and Africa often left many people stateless.\\  - Clubs: Historically, social clubs played a significant role in maintaining social hierarchies and networks.\\ Exclusion from certain clubs could have affected one's social standing and access to influential people and resources.\\- Contemporary Context:  - Statelessness: Today, statelessness remains a global issue, affecting millions of people.\\ For example, the Rohingya in Myanmar face severe statelessness and associated human rights abuses.\\  - Clubs: In modern times, while clubs still play a role in social and professional networks, the impact of exclusion is generally less severe.\\ Online communities and social media have provided alternative avenues for social interaction and networking.\\ ConclusionFine's argument that states are not clubs is supported by both points A and B.\\ Exclusion from a state has far more severe and life-altering consequences compared to exclusion from a club. \\
}

\end{tcolorbox}

\newpage

\begin{tcolorbox}[notitle,breakable,colback=white,colframe=softblue,
       boxrule=3pt, boxsep=0.5pt, enhanced, 
       shadow={3pt}{-3pt}{0pt}{opacity=1,mygrey},
       title={\large (2/4)  \includegraphics[height=22pt, trim=0 100 0 0, clip]{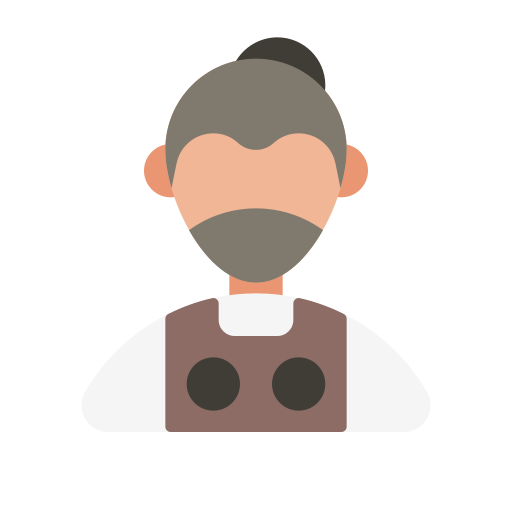}\large Historical Political Analyst}]
\texttt{
You are a Historical Political Analyst.\\ Your goal is to assess the feasibility of would-be immigrants setting up their own state, and your constraints are to focus on historical and current geopolitical realities.\\ You could follow these execution suggestions:Extract key points from the
question. Research historical attempts and current barriers to state formation by would-be immigrants.\\
\rule[0mm]{\linewidth}{0.5pt}
Fine's argument that states are not clubs is indeed grounded in the significant differences between the two in terms of the feasibility of formation and the impact of exclusion. Let's break down the options and their implications:\\ Option A: It is not possible today for would-be immigrants to get together to set up a state of their own.\\This point emphasizes the complexity and infeasibility of forming a new state. Historically, the formation of new states has been a complex and often violent process. Key factors include:\\- Territorial Control: Establishing a state requires control over a defined territory, which is often contested and requires military or political power.\\- Recognition: International recognition is crucial for a new state to be considered legitimate. Gaining such recognition involves diplomatic efforts and often the support of powerful nations.\\- Legal Frameworks: New states must establish a legal system, government institutions, and administrative structures, which are complex and resource-intensive.\\Examples of historical attempts to form new states include:\\- Kosovo: Kosovo declared independence from Serbia in 2008, but its recognition is still disputed by several countries, including Serbia and Russia.\\- South Sudan: South Sudan became independent in 2011, but the process was marked by years of civil war and ongoing ethnic conflicts.\\These examples illustrate the challenges and barriers to forming a new state, making it highly infeasible for would-be immigrants to do so.\\ Option B: Exclusion from a golf club is unlikely to have a devastating impact on the life of a would-be member.\\This point highlights the relatively minor consequences of being excluded from a club compared to being excluded from a state. Key differences include:\\- Basic Rights: Exclusion from a state can lead to statelessness, which deprives individuals of basic rights such as access to healthcare, education, and legal protection.\\- Social and Economic Impact: Statelessness can result in social isolation, economic hardship, and vulnerability to exploitation. In contrast, exclusion from a club, while potentially disappointing, does not fundamentally undermine one's ability to lead a normal life.\\ Historical and Contemporary Contexts\\- Statelessness: Historically and currently, statelessness is a significant issue. For example:\\  - Rohingya in Myanmar: The Rohingya are a stateless minority who have faced severe persecution and have been denied citizenship.\\  - Palestinian Refugees: Many Palestinians are stateless, lacking citizenship and facing significant legal and social barriers.\\- Clubs: While clubs can be important for social and professional networks, the impact of exclusion is generally less severe:\\  - Social Clubs: Exclusion from social clubs may affect one's social status and networking opportunities but does not lead to the same level of deprivation as statelessness.\\  - Professional Associations: Exclusion from professional associations can be significant for career advancement but does not deprive individuals of basic rights.\\ Conclusion\\Fine's argument that states are not clubs is supported by both points A and B. The infeasibility of forming a new state and the severe consequences of statelessness highlight the fundamental differences between states and clubs. Therefore, the correct answer is:\\Option C: Both A and B
}
\end{tcolorbox}
\newpage
\begin{tcolorbox}[notitle,breakable,colback=white,colframe=softlavender,
       boxrule=3pt, boxsep=0.5pt, enhanced, 
       shadow={3pt}{-3pt}{0pt}{opacity=1,mygrey},
       title={\large (3/4)  \includegraphics[height=22pt, trim=0 100 0 0, clip]{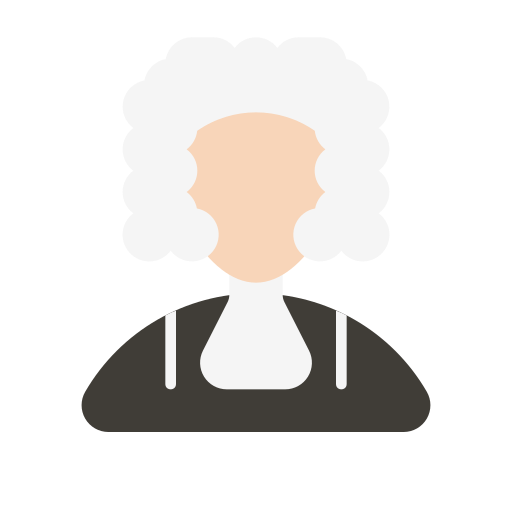}\large Legal and Institutional Framework Expert}]
\texttt{
You are a Legal and Institutional Framework Expert.\\ Your goal is to assess the legal and institutional frameworks that govern state formation and the rights of immigrants, and your constraints are to focus on both international and national legal systems.\\ You could follow these execution suggestions: Analyze the legal and institutional barriers and frameworks that affect state formation by would-be immigrants.\\
\rule[0mm]{\linewidth}{0.5pt}
Fine's argument that states are not clubs is indeed well-supported by both the infeasibility of forming a new state and the significant differences in the impact of exclusion from a state versus a club. Let's analyze the options and their implications in detail:\\ Option A: It is not possible today for would-be immigrants to get together to set up a state of their own.\\ Legal and Institutional Barriers:\\1. Sovereignty and Territorial Integrity: International law generally upholds the principle of sovereignty and territorial integrity, making it extremely difficult for groups to declare and establish new states without the consent of existing states. The United Nations Charter, for instance, emphasizes the non-interference in the internal affairs of states.\\2. Recognition: International recognition is crucial for a new state to be considered legitimate. Gaining recognition involves diplomatic negotiations and often requires the support of powerful nations. This process is complex and often politically charged.\\3. Legal Frameworks: Establishing a state requires the creation of a legal system, government institutions, and administrative structures, which are resource-intensive and require significant expertise.\\4. Resource and Military Power: Forming a new state often requires control over a defined territory, which may involve military resources and the ability to defend against external threats.\\ Historical Examples:\\- Kosovo: Kosovo declared independence from Serbia in 2008 but continues to face limited international recognition, particularly from Serbia and Russia.\\- South Sudan: South Sudan's path to independence was marked by years of civil war and ongoing ethnic conflicts, highlighting the challenges of state formation.\\ Option B: Exclusion from a golf club is unlikely to have a devastating impact on the life of a would-be member.\\ Legal and Social Impact:\\1. Basic Rights: Exclusion from a state can lead to statelessness, which deprives individuals of basic rights such as access to healthcare, education, and legal protection. Stateless individuals often face severe social and economic challenges.\\2. Social and Economic Consequences: Statelessness can result in social isolation, economic hardship, and vulnerability to exploitation. In contrast, exclusion from a club, while potentially disappointing, does not fundamentally undermine one's ability to lead a normal life.\\3. Legal Status: Statelessness can affect an individual's legal status, making them vulnerable to human rights abuses and limiting their access to essential services and opportunities.\\ Contemporary Examples:\\- Rohingya in Myanmar: The Rohingya are a stateless minority who have faced severe persecution and have been denied citizenship, leading to widespread human rights abuses.\\- Palestinian Refugees: Many Palestinians are stateless, lacking citizenship and facing significant legal and social barriers.\\ Historical and Contemporary Contexts:\\- Statelessness: Historically and currently, statelessness is a significant issue. It has been a recurring problem in various regions, such as the Middle East and Africa, often resulting from political conflicts and the redrawing of national boundaries.\\- Clubs: While clubs can be important for social and professional networks, the impact of exclusion is generally less severe. Exclusion from social clubs may affect one's social status and networking opportunities but does not lead to the same level of deprivation as statelessness.\\ Conclusion\\Fine's argument that states are not clubs is supported by both points A and B. The infeasibility of forming a new state and the severe consequences of statelessness highlight the fundamental differences between states and clubs. Therefore, the correct answer is:\\Option C: Both A and B
}
\end{tcolorbox}
\vspace{50pt}
\begin{tcolorbox}[notitle,breakable,colback=white,colframe=softcyan,
       boxrule=3pt, boxsep=0.5pt, enhanced, 
       shadow={3pt}{-3pt}{0pt}{opacity=1,mygrey},
       title={\large (4/4)  \includegraphics[height=22pt, trim=0 100 0 0, clip]{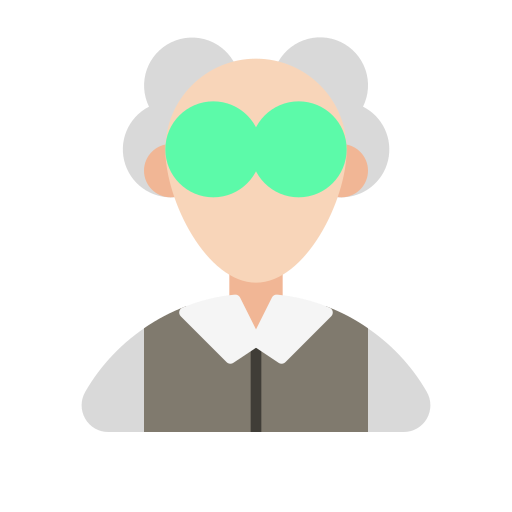}\large Marginalized Groups Analyst (Excluded)}]
\texttt{
You are a Marginalized Groups Analyst.\\ Your goal is to examine the broader context of state formation by marginalized groups, and your constraints are to consider both historical and contemporary examples.\\ You could follow these execution suggestions: Review the social and political dynamics of marginalized groups and their attempts to form states.\\
\rule[0mm]{\linewidth}{0.5pt}
None
}
\end{tcolorbox}
\vspace{50pt}
\begin{tcolorbox}[notitle,breakable,colback=white,colframe=softpeach,
       boxrule=3pt, boxsep=0.5pt, enhanced, 
       shadow={3pt}{-3pt}{0pt}{opacity=1,mygrey},
       title={\Large Final Answer}]
\texttt{
C \textcolor{green(pigment)}{\Checkmark}
}
\end{tcolorbox}

\begin{tcolorbox}[notitle, colback=red!5,colframe=softorange,
       boxrule=3pt, boxsep=0.5pt, enhanced, 
       shadow={3pt}{-3pt}{0pt}{opacity=1,mygrey},
       title={\includegraphics[height=22pt]{Fig/question-mark.png} \LARGE Query}]
\texttt{
Question: Short-term changes in plant growth rate mediated by the plant hormone auxin are hypothesized to result from
\\Option A: loss of turgor pressure in the affected cells
\\Option B: increased extensibility of the walls of affected cells
\\Option C: suppression of metabolic activity in affected cells
\\Option D: cytoskeletal rearrangements in the affected cells
}
\end{tcolorbox}
\vspace{30pt} 
\begin{tcolorbox}[notitle,breakable,colback=white,colframe=softbrown,
       boxrule=3pt, boxsep=0.5pt, enhanced, 
       shadow={3pt}{-3pt}{0pt}{opacity=1,mygrey},
       title={\large (1/5)  \includegraphics[height=22pt, trim=0 100 0 0, clip]{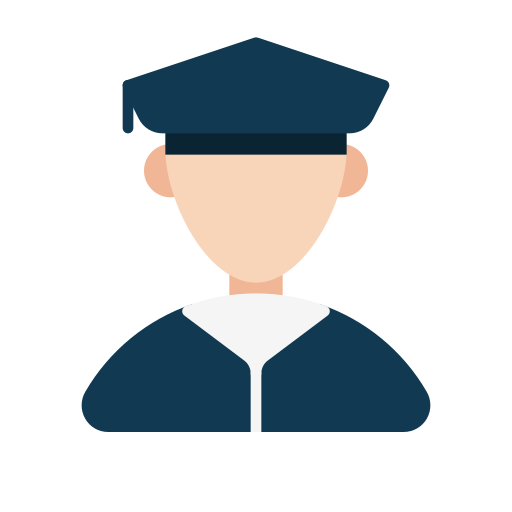}\large Auxin Physiology Specialist}]
\texttt{
You are an expert in plant physiology, named Auxin Physiology Specialist.\\Your goal is to explain the role of auxin in short-term changes in plant growth rate, and your constraints are to focus on cell elongation and growth rate.\\
\rule[0mm]{\linewidth}{0.5pt}
The correct answer isOption B: increased extensibility of the walls of affected cells.\\\\ Explanation:\\Auxin plays a critical role in regulating short-term changes in plant growth rate, primarily through its effect oncell elongation. Here’s how it works:\\\\1.Auxin and Cell Wall Extensibility: Auxin promotes the loosening of the cell wall, a process essential for cell elongation. It activates proton pumps (H$^+$-ATPases) in the plasma membrane, leading to the acidification of the cell wall (a process calledacid growth). This acidic environment activates enzymes likeexpansins, which break the hydrogen bonds between cellulose microfibrils and other cell wall components, increasing theextensibility of the cell wall.\\\\2.Turgor Pressure and Cell Elongation: With the cell wall loosened, the internal turgor pressure of the cell pushes against the wall, causing the cell to expand. This turgor-driven expansion is a key mechanism for rapid growth in plant cells.\\\\ Why Not the Other Options?\\-Option A: Loss of turgor pressure would inhibit cell elongation, not promote it. Auxin actually maintains or increases turgor pressure to drive growth.\\-Option C: Suppression of metabolic activity would slow down growth, whereas auxin stimulates metabolic processes necessary for cell expansion.\\-Option D: Cytoskeletal rearrangements are not the primary mechanism of auxin-mediated short-term growth. While the cytoskeleton is important for long-term growth and cellular organization, the immediate effect of auxin is on the cell wall.\\\\Thus, the correct answer is Option B: increased extensibility of the walls of affected cells.
}
\end{tcolorbox}

\begin{tcolorbox}[notitle,breakable,colback=white,colframe=softpink,
       boxrule=3pt, boxsep=0.5pt, enhanced, 
       shadow={3pt}{-3pt}{0pt}{opacity=1,mygrey},
       title={\large (2/5)  \includegraphics[height=22pt, trim=0 100 0 0, clip]{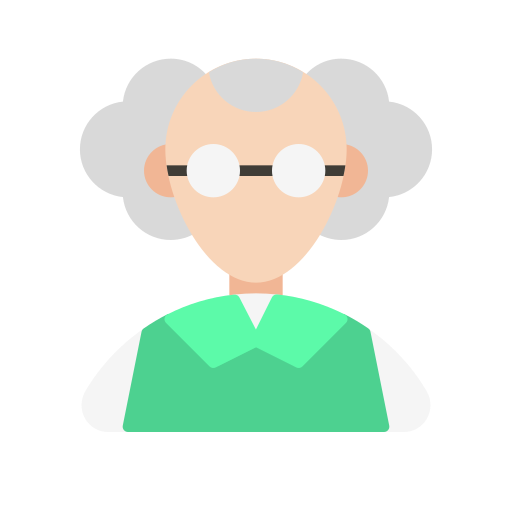}\large Molecular Mechanism Specialist (Excluded)}]
\texttt{
You are an expert in molecular biology, named Molecular Mechanism Specialist.\\ Your goal is to analyze molecular pathways and signaling processes triggered by auxin, and your constraints are to focus on gene expression changes and protein interactions.\\
\rule[0mm]{\linewidth}{0.5pt}
None
}
\end{tcolorbox}

\vspace{50pt}

\begin{tcolorbox}[notitle,breakable,colback=white,colframe=softblue,
       boxrule=3pt, boxsep=0.5pt, enhanced, 
       shadow={3pt}{-3pt}{0pt}{opacity=1,mygrey},
       title={\large (3/5)  \includegraphics[height=22pt, trim=0 100 0 0, clip]{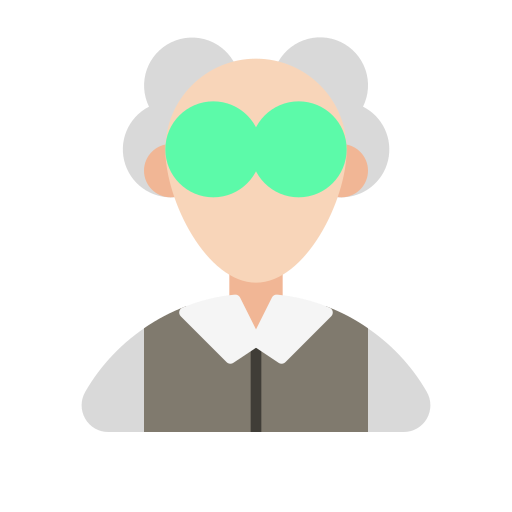}\large Enhanced Molecular Mechanism Specialist (Excluded)}]
\texttt{
 You are an expert in molecular biology, named Enhanced Molecular Mechanism Specialist.\\ Your goal is to explain the upstream molecular pathways and signaling processes by which auxin mediates short-term growth rate changes, and your constraints are to focus on gene expression, protein interactions, and how these pathways lead to specific cellular responses.\\
\rule[0mm]{\linewidth}{0.5pt}
None
}
\end{tcolorbox}
\newpage
\begin{tcolorbox}[notitle,breakable,colback=white,colframe=softlavender,
       boxrule=3pt, boxsep=0.5pt, enhanced, 
       shadow={3pt}{-3pt}{0pt}{opacity=1,mygrey},
       title={\large (4/5)  \includegraphics[height=22pt, trim=0 100 0 0, clip]{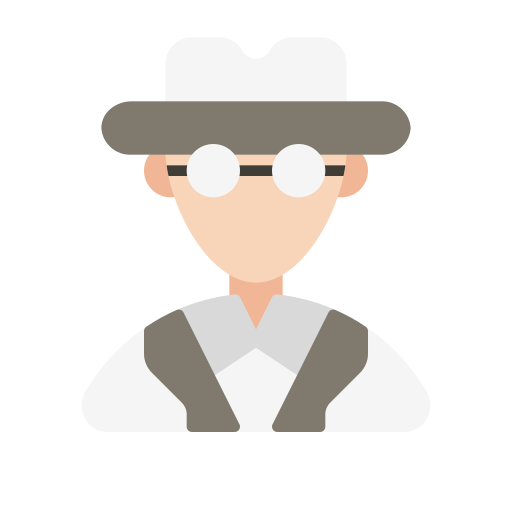}\large Cell Response Analyst}]
\texttt{
You are an expert in cellular biology, named Cell Response Analyst.\\ Your goal is to analyze cellular responses to auxin, and your constraints are to focus on changes in turgor pressure, cell wall extensibility, metabolic activity, and cytoskeletal rearrangements.\\
\rule[0mm]{\linewidth}{0.5pt}
Based on the analysis provided by Agent 7G2E (Auxin Physiology Specialist), the correct answer is indeed Option B: increased extensibility of the walls of affected cells. Here's a summary of the reasoning:\\\\ Key Points:\\1. Auxin's primary role in short-term plant growth is to promote cell elongation by increasing the extensibility of the cell wall.\\2. Mechanism: Auxin activates proton pumps (H$^+$-ATPases), leading to cell wall acidification. This acidic environment triggers the activity of enzymes like expansins, which loosen the cell wall by breaking hydrogen bonds, making the wall more extensible.\\3. Turgor Pressure: The loosened cell wall allows turgor pressure to drive cell expansion. Importantly, turgor pressure is maintained or increased by auxin, not lost (eliminating Option A).\\4. Metabolic Activity: Auxin stimulates metabolic processes required for growth, so Option C (suppression of metabolic activity) is incorrect.\\5. Cytoskeletal Rearrangements: While cytoskeletal changes are important for long-term growth and cellular organization, they are not the primary mechanism for auxin-mediated short-term growth, ruling out Option D.\\ Conclusion:The short-term changes in plant growth rate mediated by auxin are best explained by increased extensibility of the cell walls of affected cells. Therefore, the correct answer is Option B.
}
\end{tcolorbox}

\newpage
\begin{tcolorbox}[notitle,breakable,colback=white,colframe=softcyan,
       boxrule=3pt, boxsep=0.5pt, enhanced, 
       shadow={3pt}{-3pt}{0pt}{opacity=1,mygrey},
       title={\large (5/5)  \includegraphics[height=22pt, trim=0 100 0 0, clip]{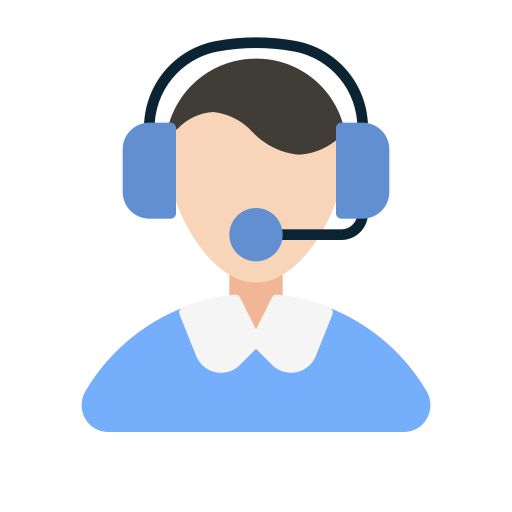}\large Hypothesis Evaluator}]
\texttt{
You are an expert in evaluating hypotheses, named Hypothesis Evaluator.\\ Your goal is to determine which option best explains the short-term changes in plant growth rate mediated by auxin, and your constraints are to use the outputs from the other roles.\\
\rule[0mm]{\linewidth}{0.5pt}
After carefully evaluating the outputs from the Auxin Physiology Specialist (Agent 7G2E) and the Cell Response Analyst (Agent wDAF), it is clear that Option B: increased extensibility of the walls of affected cells is the best explanation for short-term changes in plant growth rate mediated by auxin. Here’s the reasoning:\\1. Auxin's Primary Mechanism: Both agents highlight that auxin promotes cell elongation by increasing the extensibility of the cell wall. This is achieved through the activation of proton pumps (H+-ATPases), leading to cell wall acidification and the activation of enzymes like expansins, which loosen the cell wall.\\2. Turgor Pressure: Auxin maintains or increases turgor pressure to drive cell expansion, which rules out Option A: loss of turgor pressure.\\3. Metabolic Activity: Auxin stimulates metabolic processes necessary for growth, so Option C: suppression of metabolic activity is incorrect.\\4. Cytoskeletal Rearrangements: While cytoskeletal changes play a role in long-term growth and cellular organization, they are not the primary mechanism for auxin-mediated short-term growth, eliminating Option D.\\Both agents provide consistent and detailed evidence supporting Option B as the correct answer. The data and explanations align with established biological mechanisms of auxin action, particularly in the context of short-term growth responses.\\Final Answer: Option B: increased extensibility of the walls of affected cells.}
\end{tcolorbox}
\vspace{50pt}
\begin{figure*}[h]
    \centering
    \begin{tcolorbox}[notitle,breakable,colback=white,colframe=softpeach,
           boxrule=3pt, boxsep=0.5pt, enhanced, 
           shadow={3pt}{-3pt}{0pt}{opacity=1,mygrey},
           title={\Large Final Answer}]
    \texttt{
    B \textcolor{green(pigment)}{\Checkmark}
    }
    \end{tcolorbox}
    \caption{Case Study}
    \label{fig:case}
\end{figure*}

\begin{algorithm*}[!ht]
\DontPrintSemicolon
\SetAlgoLined
\LinesNumbered

\KwIn{Query \( q \)}
\KwOut{Final set of agents \( \mathcal{A}^* \)}

\tcp{Initialization: set previous round values to None}
$\tau^0 \gets \texttt{None}$, \( \hat{A}^0 \gets \texttt{None} \), \( \phi^0 \gets \texttt{None} \)

\For{$t \gets 1$ \KwTo $K$}{
    \tcp{Step 1: Task decomposition based on query and previous feedback}
    $\{ \tau_i^{t} \}_{i=1}^{n_t} \gets G_{p_1}(q, \phi^{t-1}, \{ \tau_i^{t-1} \}_{i=1}^{n_{t-1}}, \{ \hat{A}_i^{t-1} \}_{i=1}^{n_{t-1}})$

    \tcp{Step 2: Generate natural language agent for each sub-task}
    $\{ A_i^{t} \}_{i=1}^{n_t} \gets G_{p_2}(\{ \tau_i^{t} \}_{i=1}^{n_t})$

    \tcp{Step 3: Convert agents into standardized format}
    $\{ \hat{A}_i^{t} \}_{i=1}^{n_t} \gets G_f(\{ A_i^{t} \}_{i=1}^{n_t})$

    \If{$t < K$}{
        \tcp{Step 4: Observer provides feedback for the next round}
        $\phi^t \gets G_o(\{ \hat{A}_i^{t} \}_{i=1}^{n_t}, \{ \tau_i^{t} \}_{i=1}^{n_t})$
    }
}

\tcp{Final candidate agent set after $K$ rounds}
\( \mathcal{A}_{\text{candidate}} \gets \{ \hat{A}_i^{K} \}_{i=1}^{n_K} \)

\tcp{Generate all valid teams within the given size constraints}
\( \mathcal{T} \gets \texttt{GenerateTeams}(\mathcal{A}_{\text{candidate}}, N_{\text{min}}, N_{\text{max}}) \)

\tcp{Construct Pareto-optimal team set based on relevance and diversity}
\( \mathcal{T^*} \gets \arg \max_{\mathcal{A}' \in \mathcal{T}} \left[ \text{Rel}(\mathcal{A}', q), \text{Div}(\mathcal{A}') \right] \)

\tcp{Select final team according to task characteristics}
\( \mathcal{A}^* \gets  G_{s}(\mathcal{T^*}, q) \)

\Return {\(\mathcal{A}^*\)}

\caption{MAS Initialization Process of AgentInit}
\label{algo:full_process}
\end{algorithm*}

\begin{table*}[htb]
\centering
\scalebox{0.75}{
\begin{tabular}{ll|ccccccc}
\Xhline{1.2pt}
\textbf{Framework} & \textbf{Method}& \textbf{MMLU} & \textbf{GSM8K}&\textbf{AQUA} &\textbf{MultiArith} & \textbf{SVAMP} & \textbf{HumanEval} & \textbf{Avg.} \\
\Xhline{1.2pt}
\multicolumn{9}{c}{{Base model: Qwen2.5-72B-Instruct}} \\
\hline

\multirow{5}{*}{Chain} &\(\text{MAS}_\text{none}\)&81.0&93.3&85.0&\textbf{100.0}&91.8&85.1&89.4\\
&Pre-defined&83.7&92.6&83.6&98.8&\underline{92.9}&85.9&89.6\\
&EvoAgent&85.6&\underline{93.6}&\textbf{85.8}&99.4&91.1&84.3&90.0\\
&AutoAgents&\underline{85.7}&93.3&82.1&\textbf{100.0}&\textbf{93.4}&\underline{86.0}&\underline{90.1}\\
&AgentInit&\textbf{86.0}&\textbf{94.1}&\underline{85.0}&\textbf{100.0}&\underline{92.9}&\textbf{86.8}&\textbf{90.8}\\
\hdashline[1pt/2pt]
\multirow{5}{*}{Star} &\(\text{MAS}_\text{none}\)&82.4&93.4&83.8&\textbf{100.0}&92.5&84.3&89.4\\
&Pre-defined&84.3&\underline{93.5} &\underline{84.1}&\textbf{100.0}&\underline{92.9}&83.5&\underline{89.7}\\
&EvoAgent&\underline{85.0}&93.2&83.8&99.4&90.7&84.2&89.4\\
&AutoAgents&84.3&92.7&81.7&\textbf{100.0}&\underline{92.9}&\textbf{85.1}&89.5\\
&AgentInit&\textbf{86.9}&\textbf{93.9}&\textbf{84.6}&\textbf{100.0}&\textbf{93.0}&\textbf{85.1}&\textbf{90.6}\\
\hdashline[1pt/2pt]
\multirow{5}{*}{Layered} &\(\text{MAS}_\text{none}\)&81.7&93.4&\underline{85.0}&\textbf{100.0}&92.3&84.7&89.5\\
&Pre-defined&85.0&93.6&\underline{85.0}&\textbf{100.0}&\textbf{92.9}&85.1&\underline{90.3}\\
&EvoAgent&84.3&\textbf{94.0}&83.8&99.4&91.8&83.5&89.5\\
&AutoAgents&\underline{85.6}&93.0&82.9&\textbf{100.0}&\textbf{92.9}&\textbf{86.8}&90.2\\
&AgentInit&\textbf{86.3}&\textbf{94.0}&\textbf{85.4}&\textbf{100.0}&92.7&\underline{86.0}&\textbf{90.7}\\
\hdashline[1pt/2pt]
\multirow{5}{*}{AutoGen} &\(\text{MAS}_\text{none}\)&83.0&93.6&83.8&\textbf{100.0}&92.9&83.5&89.5\\
&Pre-defined&82.9&93.3&\underline{84.2}&\textbf{100.0}&\textbf{93.6}&\textbf{84.3}&89.7 \\
&EvoAgent &\underline{83.7}&93.9&\underline{84.2}&\textbf{100.0}&92.9&80.2&89.2\\
&AutoAgents        &83.2&\textbf{94.8}&\underline{84.2}&\textbf{100.0}&92.9&83.5&\underline{89.8}\\
&AgentInit        &\textbf{85.6}&\underline{94.1}&\textbf{84.6}&\textbf{100.0}&\underline{93.5}&\underline{83.9}&\textbf{90.3}\\
\hline
\end{tabular}}
\caption{Performance comparison between AgentInit and other MAS initialization methods across different MAS frameworks, evaluated on Qwen2.5. The best result in each framework is \textbf{bolded}, and the second-best is \underline{underlined}.}
\label{tab:qwen_main_differentF}
\end{table*}

\end{document}